\documentclass{article}
\usepackage{amsthm}
\newtheorem{theorem}{Theorem}
\usepackage{arxiv}
\usepackage[utf8]{inputenc} 
\usepackage[T1]{fontenc}    
\usepackage{hyperref}       
\usepackage{url} 
\usepackage{amssymb}
\usepackage{comment}
\usepackage{latexsym}
\usepackage{amsmath}
\usepackage{enumitem}
\usepackage{cite}
\usepackage{multirow}
\usepackage{bm}
\usepackage{makecell}

\usepackage{booktabs}       
\usepackage{amsfonts}       
\usepackage{nicefrac}       
\usepackage{microtype}      
\usepackage{lipsum}
\usepackage{graphicx}
\usepackage{subcaption}
\usepackage{xcolor}
\usepackage{algorithm}
\usepackage{algorithmic}
\graphicspath{ {./images/} }

\newcommand{\be}{\begin{eqnarray}}
\newcommand{\ee}{\end{eqnarray}}

\usepackage{amsfonts} 

\title{Complex Physics-Informed Neural Network}

\author{
  Chenhao Si \\
  School of Data Science\\
  The Chinese University of Hong Kong, Shenzhen\\
  Shenzhen, China  \\
  \texttt{222042011@link.cuhk.edu.cn} \\
  \And
  Ming Yan \\
  School of Data Science\\
  The Chinese University of Hong Kong, Shenzhen \\
  Shenzhen, China  \\
  \texttt{yanming@cuhk.edu.cn} \\
  \And
  Xin Li\\
  Department of Computer Science\\
  Northwestern University\\
  IL, USA \\
  \texttt{xinli2023@u.northwestern.edu} \\
  \And
  Zhihong Xia* \\School of Science, Great Bay University \\ Guangdong, China \\
  \& Department of Mathematics\\
  Northwestern University\\
  IL, USA \\
  \texttt{xia@math.northwestern.edu} \\
}

\begin{document}
\maketitle
\begin{abstract}
We propose compleX-PINN, a novel physics-informed neural network (PINN) architecture incorporating a learnable activation function inspired by Cauchy’s integral theorem. By optimizing the activation parameters, compleX-PINN achieves high accuracy with just a single hidden layer. Empirically, we demonstrate that compleX-PINN solves high-dimensional problems that pose significant challenges for PINNs. Our results show compleX-PINN consistently achieves substantially greater precision, often improving accuracy by an order of magnitude, on these complex tasks.
\end{abstract}


\section{Introduction}
Physics-Informed Neural Networks (PINNs) have emerged as a powerful method for solving both forward and inverse problems involving Partial Differential Equations (PDEs)~\cite{ref1, ref2, ref3, ref4}. 
PINNs leverage the expressive power of neural networks to minimize a loss function that enforces the governing PDEs and boundary/initial conditions. This approach has been widely applied across various domains, including heat transfer~\cite{ref5, ref6, ref7}, solid mechanics~\cite{ref8, ref9, ref10}, incompressible flows~\cite{ref11, ref12, ref13}, stochastic differential equations~\cite{ref14, ref15}, and uncertainty quantification~\cite{ref16, ref17}. 

Despite their success, PINNs face significant challenges and often struggle to solve certain classes of problems~\cite{ref18, ref19}. One major difficulty arises in scenarios where the solution exhibits rapid changes, such as in ‘stiff’ PDEs~\cite{ref20}, leading to issues with convergence and accuracy. To address these limitations, researchers have proposed various techniques to improve training efficiency and precision.

Over the years, numerous strategies have been developed to enhance the performance of PINNs, including adaptive weighting of loss functions and selective sampling of training points. For example, Wang et al.\cite{ref19} leveraged the Neural Tangent Kernel (NTK) to analyze gradient evolution in PINN training, adjusting the weights of each loss component accordingly. Other studies\cite{ref21, ref22} have explored methods for dynamically learning these weights during training. Additionally, adaptive sampling techniques have been introduced to tackle stiff problems by focusing on regions with high residuals. Lu et al.\cite{ref23} proposed a threshold-based approach for selecting new training points, while Wu et al.\cite{ref24} introduced a probability density function derived from residuals to improve sampling efficiency. Further extensions include active learning-based sampling methods~\cite{gao2023active} and re-sampling techniques targeting failure regions~\cite{ref26}.

Beyond these approaches, causality has been recognized as an influential factor in PINN training~\cite{ref27, RRef3}. Wang et al.~\cite{ref27} introduced a Causality-PINN, which assigns time-dependent weights to the loss function for time-dependent PDEs, while Daw et al.\cite{RRef3} further integrated causality with importance sampling techniques. 

Recent advancements in Physics-Informed Neural Networks (PINNs) include domain decomposition methods~\cite{ref28, ref29, ref30, ref31}, improved initialization schemes~\cite{ref31, ref32, ref33}, novel loss functions~\cite{RRef5}, and innovative network architectures. For example, Kolmogorov-Arnold Networks (KANs)~\cite{liu2024kan}—originally designed for general function representation—employ spline-based activation functions inspired by the Kolmogorov-Arnold theorem to enhance expressivity. Here, we adapt KANs for solving partial differential equations (PDEs), termed PIKAN~\cite{shukla2024comprehensive,wang2025kolmogorov}, to leverage their adaptive basis functions while addressing challenges in high-dimensional scalability. Alongside other architectural innovations like hierarchical networks~\cite{RRef2, PIRBN, ref34, ref35,wang2024piratenets}, such advancements—including our PDE-focused adaptation of KANs—demonstrate the field’s progress in tailoring neural networks to the structure of physical systems. Collectively, these improvements have expanded PINNs’ utility for complex real-world applications, though optimizing expressivity and efficiency in high dimensions remains critical.

However, existing methods primarily rely on additional training techniques or significantly larger models, such as transformers~\cite{RRef2} and convolutional neural networks~\cite{ref35}. While these approaches can enhance performance, they also introduce substantial computational overhead. Auxiliary networks and gradient-based modifications~\cite{ref36, ref37} increase training costs, whereas larger models suffer from slower convergence due to the large number of parameters~\cite{ref18, RRef4}. Even recent innovations like KANs, despite their theoretical promise, face scalability challenges in high-dimensional PDE settings, as shown in our experiments (Section~\ref{Section Poisson}).

To overcome these limitations, developing a network that enhances both convergence and predictive accuracy without relying on additional training algorithms or significantly increasing model complexity is crucial. Motivated by this, we introduce compleX-PINN, which incorporates the Cauchy integral formula~\cite{XNet1, XNet2} into a novel activation function, offering a more efficient and effective alternative to traditional PINNs.

The main contributions of this paper are summarized as follows: 
\begin{itemize}
    \item To the best of our knowledge, this is the first comprehensive study on using Cauchy-based activation functions in PINNs.
    \item We provide a detailed derivation and motivation for incorporating Cauchy activation functions into PINN architectures.
    \item Empirical results demonstrate that compleX-PINN outperforms several PINN-based models.
    \item We show that compleX-PINN is compatible with existing PINN training techniques, further enhancing its performance when integrated with these methods.
\end{itemize}

The organization of this paper is as follows. Section~\ref{Sec2.1} provides a brief introduction to PINNs. Our proposed model, compleX-PINN, is introduced in Section~\ref{Cauchy activation function}, where we first present the Cauchy activation function using Cauchy's 1D integral formula in Section~\ref{Sec 3.1}, extend it to high-dimensional cases in Section~\ref{Sec 3.2}, and apply it to neural networks in Section~\ref{Extend to high dimension}.  
Finally, numerical results for testing several PDEs in high dimension are presented in Section~\ref{Sec4}, followed by the conclusion.

\section{Physics-Informed Neural Network}
\label{Sec2.1}
Denote the spatial domain as $\Omega \subset \mathbb{R}^n$ with boundary $\partial \Omega$, and let $T$ represent the time domain. The spatial-temporal variable is given by $(\mathbf{x}, t) \in \Omega \times T$. A time-dependent partial differential equation (PDE) over this domain is defined as follows:
\begin{align}
    \mathcal{F}[u](\mathbf{x}, t) &= 0, \label{(1)} \quad (\mathbf{x}, t) \in \Omega \times T, \\
    \mathcal{B}[u](\mathbf{x}, t) &= 0, \label{(2)} \quad (\mathbf{x}, t) \in \partial \Omega \times T, \quad \text{(boundary condition)} \\
    \mathcal{I}[u](\mathbf{x}, 0) &= 0, \quad \mathbf{x} \in \Omega, \hspace{51pt} \text{(initial condition)}
\end{align}
where $\mathcal{F}$, $\mathcal{B}$, and $\mathcal{I}$ are differential operators, and $u(\mathbf{x}, t)$ is the solution to the PDE, subject to boundary and initial conditions.

A PINN parameterized by $\theta$ approximates the solution $u(\mathbf{x}, t)$. The input to the neural network is $(\mathbf{x}, t)$, and the approximation is denoted by $\hat{u}(\theta)(\mathbf{x}, t)$. The PINN minimizes the following objective function:
\begin{align}
    \mathcal{L}(\theta) = \lambda_F \mathcal{L}_F(\theta) + \lambda_B \mathcal{L}_B(\theta) + \lambda_I \mathcal{L}_I(\theta), \label{(3)}
\end{align}
where
\begin{align}
    \mathcal{L}_F(\theta) &= \frac{1}{N_f} \sum_{(\mathbf{x}, t) \in \Omega_F} \big| \mathcal{F}[\hat{u}(\theta)](\mathbf{x}, t) \big|^2, \label{(4)} \\
    \mathcal{L}_B(\theta) &= \frac{1}{N_b} \sum_{(\mathbf{x}, t) \in \Omega_B} \big| \mathcal{B}[\hat{u}(\theta)](\mathbf{x}, t) \big|^2, \label{(5)} \\
    \mathcal{L}_I(\theta) &= \frac{1}{N_0} \sum_{(\mathbf{x}, 0) \in \Omega_I} \big| \mathcal{I}[\hat{u}(\theta)](\mathbf{x}, 0) \big|^2. \label{(6)}
\end{align}
Here, $\Omega_F$, $\Omega_B$, and $\Omega_I$ are the training sets for the PDE residual, boundary condition, and initial condition, respectively, with cardinalities $N_f$, $N_b$, and $N_0$. The weights $\lambda_F$, $\lambda_B$, and $\lambda_I$ are hyper-parameters tuning the contributions of each loss component. Notably, $\Omega_F$ may include points on the boundary or at the initial time, allowing $\Omega_F \cap \Omega_B$ and $\Omega_F \cap \Omega_I$ to be non-empty.

The choice of activation function $\sigma(\cdot)$ is crucial in PINNs, as it introduces the nonlinearity necessary to approximate complex solutions to PDEs. The hyperbolic tangent (tanh) is commonly used in PINNs for its smoothness and training stability~\cite{ref1, ref2}. Wavelet-based activations have also been explored to capture multi-scale features~\cite{ref40, RRef2, ref34}. Recent work by Li et al. \cite{XNet1} proposed the Cauchy activation function, which has shown strong performance in computer vision and time-series forecasting tasks \cite{XNet2}. 

In this study, we extend the application of the Cauchy activation function to address PDEs that are challenging for standard PINNs. The Cauchy activation function and the novel compleX-PINN model are introduced in the following section.

\section{Complex Physics-Informed Neural Network}
\label{Cauchy activation function}

The Cauchy activation function, introduced in \cite{XNet1}, is defined as:
\begin{align}
    \Phi(x; \mu_1, \mu_2, d) = \frac{\mu_1 x}{x^2 + d^2} + \frac{\mu_2}{x^2 + d^2}, \label{(12)}
\end{align}
where $\mu_1$, $\mu_2$, and $d$ are trainable parameters. Moreover, $\mu_1$ governs the linear component of the input $x$, $\mu_2$ regulates the constant, enhancing adaptability in shaping the activation, and $d$ controls the activation function’s range and smoothness properties. The initialization effect of each parameters will be discussed later in Section~\ref{Sec Robustness}. This activation function is inspired by Cauchy's integral formula, as we further elaborate in Section~\ref{Sec 3.1}. We refer to a PINN model employing the Cauchy activation function as compleX-PINN.

We would like to note at the outset that our network is initially constructed with a single hidden layer, where each neuron has a unique set of trainable parameters $\{\mu_1,\mu_2,d\}$. Consequently, the total number of trainable parameters for the Cauchy activation function is $3\times N_{\text{Cauchy}}$, where $N_{\text{Cauchy}}$ represents the width of the layer.

\subsection{1D Cauchy's integral formula and the Cauchy activation function}
\label{Sec 3.1}

This section introduces Cauchy's integral formula and derives the Cauchy activation function from it.

\begin{theorem}[Cauchy's Integral Formula]
\label{theorem 1}
Let $f$ be a complex-valued function on the complex plane. If $f$ is holomorphic inside and on a simple closed curve \( C \), and $z$ is a point inside \( C \), then:
\[
f(z) = \frac{1}{2\pi i} \oint_C \frac{f(\zeta)}{\zeta - z} \, d\zeta.
\]
\end{theorem}

Cauchy's integral formula expresses the value of a function at any point $z$ as a function of known values along a closed curve \( C \) that encloses \( z \). 
Remarkably, this principle is akin to machine learning, where the values at new points are inferred from the known values.

In practice, we approximate the integral using a Riemann sum over a finite number of points on the curve \( C \). Let \( \zeta^1, \zeta^2, \ldots, \zeta^m \) be a sequence of \( m \) points on \( C \). Then,
\begin{align} 
    f(z) \approx \frac{1}{2\pi i} \sum_{k=1}^{m} \frac{f(\zeta^{k})}{\zeta^{k} - z} \, (\zeta^{k+1} - \zeta^{k}) := \sum_{k=1}^m \frac{\lambda_k}{\zeta^k - z}, \label{equ9}
\end{align}
where, for convenience, we set \( \zeta^{m+1} = \zeta^1 \) and define \( \lambda_k = \frac{f(\zeta^k) \, (\zeta^{k+1} - \zeta^k)}{2\pi i} \).

If our target function \( f \) is real and one-dimensional, we obtain:
\begin{align}
    f(x) \approx \text{Re} \left( \sum_{k=1}^m \frac{\lambda_k}{\zeta^k - x} \right) = \sum_{k=1}^m \frac{\text{Re}(\lambda_k) \, \text{Re}(\zeta^k) + \text{Im}(\lambda_k) \, \text{Im}(\zeta^k) - \text{Re}(\lambda_k) x}{(x - \text{Re}(\zeta^k))^2 + (\text{Im}(\zeta^k))^2}. \label{(11)}
\end{align}

With the Cauchy activation function defined in~\eqref{(12)}, we have
\begin{align}
    f(x) \approx \sum_{k=1}^m \Phi \left( x - \text{Re}(\zeta^k); -\text{Re}(\lambda_k), \, \text{Re}(\lambda_k) \, \text{Re}(\zeta^k) + \text{Im}(\lambda_k) \, \text{Im}(\zeta^k), \, (\text{Im}(\zeta^k))^2 \right).
\end{align}
This shows that a one-layer neural network with the Cauchy activation function~\eqref{(12)} can approximate the real function \( f(x) \). 


At this stage, we emphasize the unique strength of the Cauchy activation function embedded in compleX-PINN. Unlike traditional activation functions, such as ReLU or Sigmoid, which typically offer only first-order approximation capabilities and are thus limited in expressiveness. The Cauchy activation function provides a natural path to high-order and even exponential approximation accuracy. 

The key idea is that the activation function is derived from a discretized version of the Cauchy integral formula, where the network learns both the sample points and the associated weights. For analytic functions defined on compact domains, the approximation error of the Cauchy integral formula decays as $O(r^{-m})$ for some $r < 1$, where $m$ is the number of sampling points and $r$ depends on the analyticity properties of the function and the geometry of its domain~\cite{trefethen2014exponentially}. This exponential convergence is significantly faster than the algebraic decay $O(m^{-p})$ achieved by polynomial-based approximations of order $p$. In other words, the Cauchy activation function enables more efficient and accurate function representation, particularly for smooth or analytic functions.

While various numerical integration schemes—such as the trapezoidal rule, Simpson’s rule, Boole’s rule, or more general Newton–Cotes formulas—can be used to discretize the Cauchy integral, they ultimately produce the same functional form for the activation, up to the specific values of the quadrature weights $\lambda_k$ and nodes $z_k$. Since these quantities are treated as trainable parameters in the neural network, the choice of integration rule becomes inconsequential. The network adapts them to best approximate the target function during training.



\subsection{Multi-dimensional Cauchy's integral formula}
\label{Sec 3.2}

This section extends Cauchy's integral formula to the multi-dimensional case. 

\begin{theorem}[Multi-Dimensional Cauchy's Integral Formula]
\label{theorem 2}
Let \( f(z) \) be holomorphic in a compact domain \( U \subset \mathbb{C}^N \) within \( N \)-dimensional complex space. For simplicity, assume that \( U \) has a product structure: \( U = U_1 \times U_2 \times \ldots \times U_N \), where each \( U_i \), \( i = 1, 2, \dots, N \), is a compact domain in the complex plane. Let \( P \) denote the surface defined by
\[
P = \partial U_1 \times \partial U_2 \times \ldots \times \partial U_N,
\]
then a multi-dimensional extension of Cauchy's integral formula for \( (z_1, z_2, \ldots, z_N) \in U \) is given by:
\[
f(z_1, z_2, \ldots, z_N) = \left(\frac{1}{2\pi i}\right)^N \int\cdots\int_P \frac{f(\zeta_1, \zeta_2, \ldots, \zeta_N)}{(\zeta_1 - z_1)(\zeta_2 - z_2) \cdots (\zeta_N - z_N)} \, d\zeta_1 \cdots d\zeta_N.
\]
\end{theorem}

Similarly, we approximate the integral by a Riemann sum over a finite number of points. More precisely, for any integer \( l = 1, \ldots, N \), let
\( \zeta_l^1, \zeta_l^2, \ldots, \zeta_l^{m_l} \) be a sequence of \( m_l \) points on \( \partial U_l \). Then,
\begin{align}
    f(z_1, z_2, \ldots, z_N) \approx \left(\frac{1}{2\pi i}\right)^N
    \sum_{k_1=1}^{m_1} \cdots \sum_{k_N=1}^{m_N}
    \frac{f(\zeta_1^{k_1}, \zeta_2^{k_2}, \ldots, \zeta_N^{k_N})}{(\zeta_1^{k_1} - z_1)(\zeta_2^{k_2} - z_2) \cdots (\zeta_N^{k_N} - z_N)}
    (\zeta_1^{k_1+1} - \zeta_1^{k_1}) \cdots (\zeta_N^{k_N+1} - \zeta_N^{k_N}),
\end{align}
where, for convenience, we set \( \zeta_l^{m_l+1} = \zeta_l^1 \) for \( l = 1, 2, \ldots, N \).

Collecting all terms that are independent of \( z_1, \ldots, z_N \), we define
\[
\lambda_{k_1, \ldots, k_N} = \left(\frac{1}{2\pi i}\right)^N f(\zeta_1^{k_1}, \zeta_2^{k_2}, \ldots, \zeta_N^{k_N}) (\zeta_1^{k_1+1} - \zeta_1^{k_1}) \cdots (\zeta_N^{k_N+1} - \zeta_N^{k_N}),
\]
so that we can rewrite the approximation as
\begin{align}
    f(z_1, z_2, \ldots, z_N) \approx \sum_{k_1=1}^{m_1} \cdots \sum_{k_N=1}^{m_N} \frac{\lambda_{k_1, \ldots, k_N}}{(\zeta_1^{k_1} - z_1)(\zeta_2^{k_2} - z_2) \cdots (\zeta_N^{k_N} - z_N)}.
\end{align}

Since the order of the sample points no longer matters, we can rewrite the sample points as a single sequence \( (\zeta_1^k, \ldots, \zeta_N^k) \) for \( k = 1, 2, \ldots, m \), where \( m = m_1 m_2 \cdots m_N \). Thus, we finally obtain
\begin{align}
    f(z_1, z_2, \ldots, z_N) \approx \sum_{k=1}^m \frac{\lambda_k}{(\zeta_1^k - z_1)(\zeta_2^k - z_2) \cdots (\zeta_N^k - z_N)}, \label{(8)}
\end{align}
where \( \lambda_1, \lambda_2, \ldots, \lambda_m \) are parameters that depend on the sample points \( (\zeta_1^k, \zeta_2^k, \ldots, \zeta_N^k) \) and the values \( f(\zeta_1^k, \zeta_2^k, \ldots, \zeta_N^k) \) for \( k = 1, 2, \ldots, m \).

While the multi-dimensional Cauchy integral formula provides a powerful theoretical foundation for approximating holomorphic functions, its direct application in high-dimensional settings poses computational challenges. Specifically, the number of terms in the approximation grows exponentially with dimension $N$, and the resulting structure can be difficult to implement and optimize efficiently within standard neural network architectures. To address this, we propose a more practical alternative in the next section: approximating high-dimensional functions by applying the Cauchy activation function to suitable linear projections of the input. This approach not only retains the analytic expressiveness of the Cauchy activation function but also offers scalability and adaptability in learning complex features from high-dimensional data.


\subsection{Using Cauchy activation to approximate high dimensional functions}
\label{Extend to high dimension}
The Cauchy approximation formula derived above can be computationally inefficient when the dimension $N$ is high, due to the large number of multiplicative terms in the denominator. Therefore, for large $N$, we opt for a simplified representation of the function by applying the Cauchy activation function to linear combinations of the variables. Generally, this corresponds to a dual representation of the function, which is especially efficient for feature-finding in high-dimensional problems.  Specifically, we approximate the target function \( f(x_1, x_2, \dots, x_N) \) by
\begin{align}
    f(x_1, x_2, \dots, x_N) \approx \sum_{k=1}^m \Phi(W_{k1} x_1 + W_{k2} x_2 + \dots + W_{kN} x_N + b_k; \mu_{k1}, \mu_{k2}, d_k), \label{eq:multi_cauchy_activation}
\end{align}
where each \( \Phi \) is a Cauchy activation function as defined in Equation~\eqref{(12)}. Here, the parameters $W_{k1}, W_{k2}, \ldots, W_{kN}$, $b_k, \mu_{k1}, \mu_{k2}$, and \( d_k \) are trainable, allowing the network to capture the complex relationships among the input variables. The validity of this approximation structure is grounded in the following theorem~\cite{XNet1}:

\begin{theorem}[General Approximation Theorem] \label{thm:general-approximation}
Let \(\Phi\) be a family of functions in \(\mathcal{C}(\mathbb{R}, \mathbb{R})\), the space of continuous functions from $\mathbb{R}$ to $\mathbb{R}$, with the universal approximation property. Define
\[
\Phi^N = \left\{ \phi_a(a_1 x_1 + a_2 x_2 + \cdots + a_N x_N) \mid (a_1, \ldots, a_N) \in \mathbb{R}^N, \phi_a \in \Phi \right\}.
\]
Then, the family \(\Phi^N\) has the universal approximation property in \(\mathcal{C}(\mathbb{R}^N, \mathbb{R})\), i.e., every continuous function in \(\mathbb{R}^N\) can be approximated by a linear combination of functions in \(\Phi^N\), uniformly over compact subsets.
\end{theorem}

Since the Cauchy activation function $\Phi$ possesses the universal approximation property in one dimension (as shown in Equation~\eqref{(12)}), Theorem~\ref{thm:general-approximation} guarantees that the family in $\Phi^N$ \eqref{eq:multi_cauchy_activation} can approximate any continuous function in $\mathbb{R}^N$, uniformly over compact subsets. For completeness, we outline the proofs of this theorem based on Ref.~\cite{XNet1}.

\begin{proof}
The set of all polynomials is dense in $\mathcal{C}(B,\mathbb{R})$ for any compact subset $B\subset \mathbb{R}^N$, by the Stone–Weierstrass Theorem. Since an analytical function $f$ on a compact domain can be approximated arbitrarily well by a polynomial, it suffices to show that each monomial can be approximated, up to an arbitrary order, by the Cauchy activation function.

As demonstrated earlier, the activation function $\Phi$ is capable of approximating one-dimensional monomials
$f(x) = x^k$ for any integer $k$. This immediately implies that $\Phi$, when applied to linear combinations in $\mathbb{R}^N$, can approximate functions of the form
$$
(a_1x_1 + a_2x_2 + \ldots + a_Nx_N)^k, \; \mbox{ for } k\in \mathbb{N},\;
(a_1, \ldots, a_N) \in \mathbb{R}^N.
$$ 
Fix an integer $k$, the multinomial expansion of the above function yields a linear combination of monomials:
$$x_1^{k_1} x_2^{k_2} \cdots x_N^{k_N}, \; \; k_1 + k_2 + \ldots +k_N
= k.$$
Conversely, by the Waring decomposition~\cite{ranestad2000varieties}, any such multivariate monomial can itself be expressed as a linear combination of functions of the form 
$$
(a_1x_1 + a_2x_2 + \ldots + a_Nx_N)^k.
$$
Hence, the Cauchy activation function is capable of approximating any polynomial function and therefore any analytic function on compact domains in $\mathbb{R}^N$, establishing its expressive power in high dimensions.
\end{proof}

Cauchy activation offers a uniquely expressive framework for approximating complicated functions. A single layer with Cauchy activation achieves high precision in practice, often eliminating the need for multiple activation layers. While other architectural components (e.g., LSTMs) may still be necessary for domain-specific tasks~\cite{XNet2}, the simplicity of this design reduces computational overhead while maintaining accuracy.

\section{Numerical Experiments}
\label{Sec4}
CompleX-PINN is uniquely characterized by a single hidden layer, referred to as the \textbf{Cauchy layer}. Unlike other neural networks, each neuron in the Cauchy layer is parameterized by trainable parameters $\{\mu_1, \mu_2, d\}$, which provide greater flexibility in capturing complex solution patterns. By default, the parameters $\{\mu_1, \mu_2, d\}$ are initialized to $0.1$, unless otherwise specified in particular experiments. Denote the number of neurons in the Cauchy layer as $N_{\text{Cauchy}}$, then the total number of parameters for a $D$-dimensional PDE with the 1-dimensional output is \[(D+1+3)\times N_{\text{Cauchy}} + (N_{\text{Cauchy}} + 1)\times1.\]

Although each neuron introduces three additional parameters, the overall number of trainable parameters in compleX-PINN remains significantly smaller than currently used multi-layer architectures, as a single Cauchy layer is sufficient. 

CompleX-PINN is evaluated against three state-of-the-art models representing distinct technical approaches: 
\begin{itemize}
    \item Residual-Based Attention PINN (RBA-PINN)~\cite{RRef4}. This method adaptively adjusts the residual loss weights at individual training points to emphasize more critical regions in the solution domain. It employs a standard fully connected neural network with depth $D_{\text{RBA}}$ and width $N_{\text{RBA}}$.  The total number of trainable parameters is given by  
    \[(D+1) \times N_{\text{RBA}} + \sum_{i=1}^{D_{\text{RBA}} - 1}(N_{\text{RBA}}+1) \times N_{\text{RBA}}  + (N_{\text{RBA}} + 1)\times 1.\]

    \item Binary structured PINN (BsPINN)~\cite{ref34}. This method employs a hierarchical binary architecture that systematically reduces inter-neuron connections, enhancing computational efficiency while maintaining expressive power. The network configuration is denoted as $M$-\(\tilde{M}\), where the first hidden layer contains $M$ neurons grouped in a single block, and each subsequent layer halves the block size until it reaches \(\tilde{M}\). This structure leads to $D_{\text{BsPINN}}:=\log_2(M/\tilde{M})+1$ hidden layers. The total number of parameters is calculated as 
    \[(D+1) \times M + \sum_{i=1}^{D_{\text{BsPINN}}-1} (2^{1-i}M+1)\times M  + (M+1)\times 1.\]
    
    \item Physics-Informed Kolmogorov-Arnold Network (PIKAN)~\cite{liu2024kan,shukla2024comprehensive,wang2025kolmogorov}. PIKAN\footnote{The code of KAN is adapted from the efficientKAN \href{https://github.com/Blealtan/efficient-kan}{https://github.com/Blealtan/efficient-kan}.} leverages the Kolmogorov-Arnold representation theorem by replacing traditional linear transformations with univariate B-spline functions. Given a spline order $p$ and $G$ intervals (which is equivalent to $G+1$ grid points), each univariate function is represented by $G+p$ basis functions. Each layer includes two additional learnable weight components—both initialized to $1$~\cite{liu2024kan}—corresponding to the base function and the spline function. These components contribute to the overall parameter count. With network width $N_{\text{KAN}}$ and depth $D_{\text{KAN}}$, as indicated in Ref.~\cite{wang2025kolmogorov}, the total number of parameters is 
    \[D\times N_{\text{KAN}}\times(G + p +2) + \sum_{i=1}^{D_{\text{KAN}} - 1} N_{\text{KAN}}\times N_{\text{KAN}}\times(G + p +2) + N_{\text{KAN}}\times 1\times(G + p+2).\]
\end{itemize}

The experimental setup is organized into distinct subsections, each targeting specific objectives, as outlined below:

\begin{itemize}
    \item \textbf{Section~\ref{Sec Helm}}: 
    This section focuses on 2D Helmholtz problems, for which all three models have published results (albeit with varying PDE parameters). To ensure a fair and consistent comparison, we directly adopt the reported results from the original references, using identical training settings for each model. This avoids potential discrepancies due to reimplementation and preserves alignment with established baselines.
    
    \item \textbf{Section~\ref{Sec 3D Helmholtz}-\ref{Section Poisson} (high-dimensional PDEs)}: In these sections, we explore high-dimensional PDEs by reproducing state-of-the-art methods using their publicly available codebases, adapted to our specific target equations. To ensure fair comparisons, we adhere to a standardized evaluation protocol across all methods. Moreover, to mitigate biases due to hyperparameter sensitivity and randomness in training, we perform extensive hyperparameter tuning for each model:
    \begin{itemize}
        \item \textbf{RBA-PINN}: depth $D_{\text{RBA}}\in\{3, 4, 5\}$; width  $N_{\text{RBA}}\in\{50, 80, 120, 200\}$.
        \item \textbf{BsPINN}: block size configurations: $\{128-8, 128-16, 256-16, 256-32, 512-32, 512-64\}$.
        \item \textbf{PIKAN}: number of intervals $G \in \{2, 4, 9\}$; width $N_{\text{KAN}} \in \{5, 10, 15\}$, with depth fixed as $D_{\text{KAN}} = 2$ and spline order $ p =3$, in accordance with common configurations used in prior works~\cite{shukla2024comprehensive,wang2025kolmogorov}
    \end{itemize}
    Following the tuning process, we report the experimental results corresponding to the best-performing configurations, selected based on validation performance.
\end{itemize}



    


{\bf Training and testing points.} For 2D spatial or 1D spatio-temporal PDEs, test points are generated on a uniform $300\times 300$ grids (90,000 points). For higher-dimensional problems, 90,000 test points are randomly sampled to balance computational and memory constraints. Training points are also randomly selected, with $N_f$, $N_b$, and $N_0$ denoting the numbers of residual, boundary, and initial condition points, respectively. Note that training points are not necessarily included in the testing set.

For each PDE, each model is trained and evaluated across three independent trials using different random seeds. The best-performing result among the trials is reported. Performance is measured using the relative $L^2$ error and the $L^{\infty}$ norm, defined as follows:
\begin{align}
    \text{(Relative) } L^2 \text{ error} & = \frac{\sqrt{\sum_{k=1}^N|\hat{u}(\mathbf{x}_k, t_k) - u(\mathbf{x}_k, t_k)|^2}}{\sqrt{\sum_{k=1}^N |u(\mathbf{x}_k, t_k)|^2}},
\label{L2error}\\
    L^{\infty} \text{ norm} & = \max_{1 \leq k \leq N} \left|\hat{u}(\mathbf{x}_k, t_k) - u(\mathbf{x}_k, t_k)\right|,\label{Linferror}
\end{align}
where $u$ denotes the ground truth solution, $\hat{u}$ is the prediction from the tested model, and $N$ is the number of testing points.

\subsection{2D Helmholtz equation}
\label{Sec Helm}

The Helmholtz equation, widely used to model wave propagation and diffusion phenomena, describes physical processes that evolve over spatial or spatio-temporal domains. The two-dimensional (2D) Helmholtz equation has been extensively studied in the literature \cite{RRef4, ref21, ref22, ref36, son2023enhanced, ref18, shukla2024comprehensive, ref34}. Therefore, we compare compleX-PINN directly with existing results using the same problem setups reported in these studies.

We consider the following form of the 2D Helmholtz equation:
\begin{align}
    &u_{xx}+u_{yy} + k^2u - q(x,y) = 0, \quad (x,y) \in \Omega, \\
    &u(x,y) = 0, \quad (x,y)\in\partial\Omega, 
\end{align}
where the source term $q(x,y)$ is defined as:
\begin{align}
    q(x,y) = k^2\sin(a_1\pi x)\sin(a_2\pi y)-(a_1\pi)^2\sin(a_1\pi x)\sin(a_2\pi y)-(a_2\pi)^2\sin(a_1\pi x)\sin(a_2\pi y).
\end{align}
We set $k = 1$ and $\Omega = [-1, 1]\times [-1,1]$. The exact solution to the equation is given by
\begin{align}
    u(x,y) = \sin(a_1\pi x)\sin(a_2\pi y),
\end{align}
where the parameters $a_1$ and $a_2$ control the spatial frequency components of the solution.

\subsubsection{Comparison with baseline models}
We evaluate compleX-PINN against several baseline models: RBA-PINN~\cite{RRef4}, PIKAN~\cite{shukla2024comprehensive,wang2025kolmogorov}(with $a_1 = 1$ and $a_2 = 4$), and BsPINN~\cite{ref34} (with $a_1 = a_2 = 8$). Each model is tested under its original PDE configuration, and we strictly follow the training protocols specified in the respective studies. This approach ensures fair, architecture-focused comparisons by eliminating confounding factors such as differences in hyperparameter tuning.

\paragraph{RBA-PINN Comparison (Case 4.1.1a):}\label{sec:411a} We construct a compleX-PINN model with 500 neurons in the Cauchy layer. For comparison, the RBA-PINN employs a 5-layer fully connected neural network with 128 neurons per layer. To enforce boundary conditions, we use a hard constraint formulation:
\[\hat{u} = (1-x^2)(1-y^2)\hat{u}_{\mathcal{NN}}\]
where $\hat{u}_{\mathcal{NN}}$ is the raw network output and $\hat{u}$ is the final prediction satisfying the PDE boundary conditions.

To ensure comparability, we adopt the same number of training points $N_f = 256,000$ and the same number of iterations as in RBA-PINN~\cite{RRef4}. Specifically, we train using the Adam optimizer for 20,000 iterations with a learning rate of 0.005, followed by 1,000 iterations of L-BFGS for fine-tuning. An exponential learning rate scheduler with decay rate 0.7 is applied every 1,000 steps. All training details—including learning rates, scheduling, and network structure—closely follow the publicly available code provided by RBA-PINN.



\begin{figure}[!htb]
    \centering
    \begin{minipage}{0.6\textwidth}
    \centering
    \includegraphics[width=\linewidth]{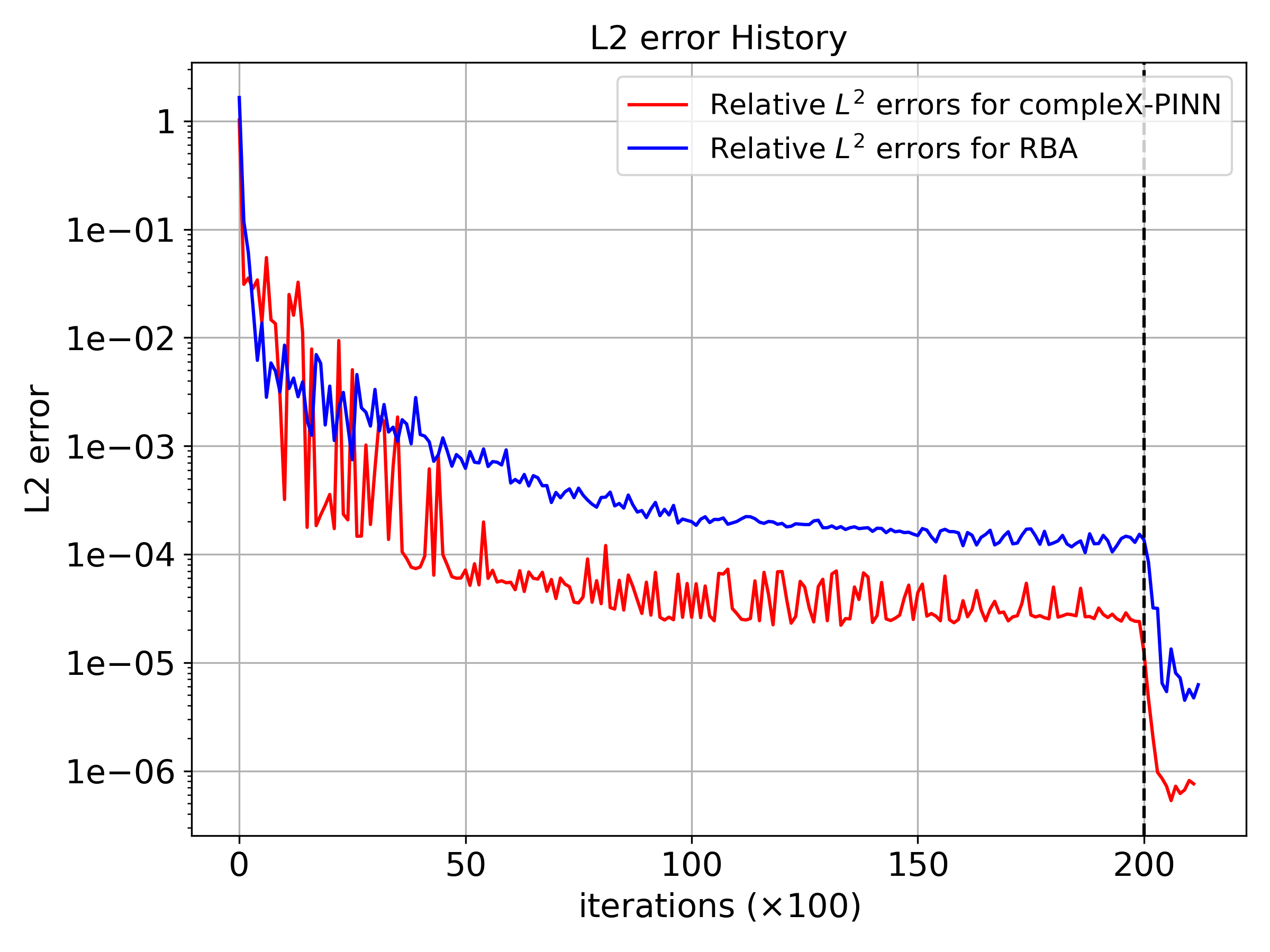}    
    \end{minipage}

    \caption{Relative $L^2$ error history of the 2D Helmholtz equation using the Adam optimizer (red: compleX-PINN; blue: RBA-PINN), followed by 1,000 iterations of the L-BFGS optimizer. A dashed vertical line indicates the transition from Adam to L-BFGS. The Adam optimizer is configured with a learning rate of 0.005 and an exponential decay scheduler (decay rate: 0.7; decay step: 1,000). Although the large initial learning rate causes some oscillation in the first 5,000 iterations, the relative  $L^2$ error stabilizes as the learning rate decays. The subsequent L-BFGS phase further refines the solution.}\label{Helmholtz L2 HIST}
\end{figure}

\begin{figure}[!htb]
    \centering
    \begin{minipage}{0.45\textwidth}
    \centering
    \includegraphics[width=\linewidth]{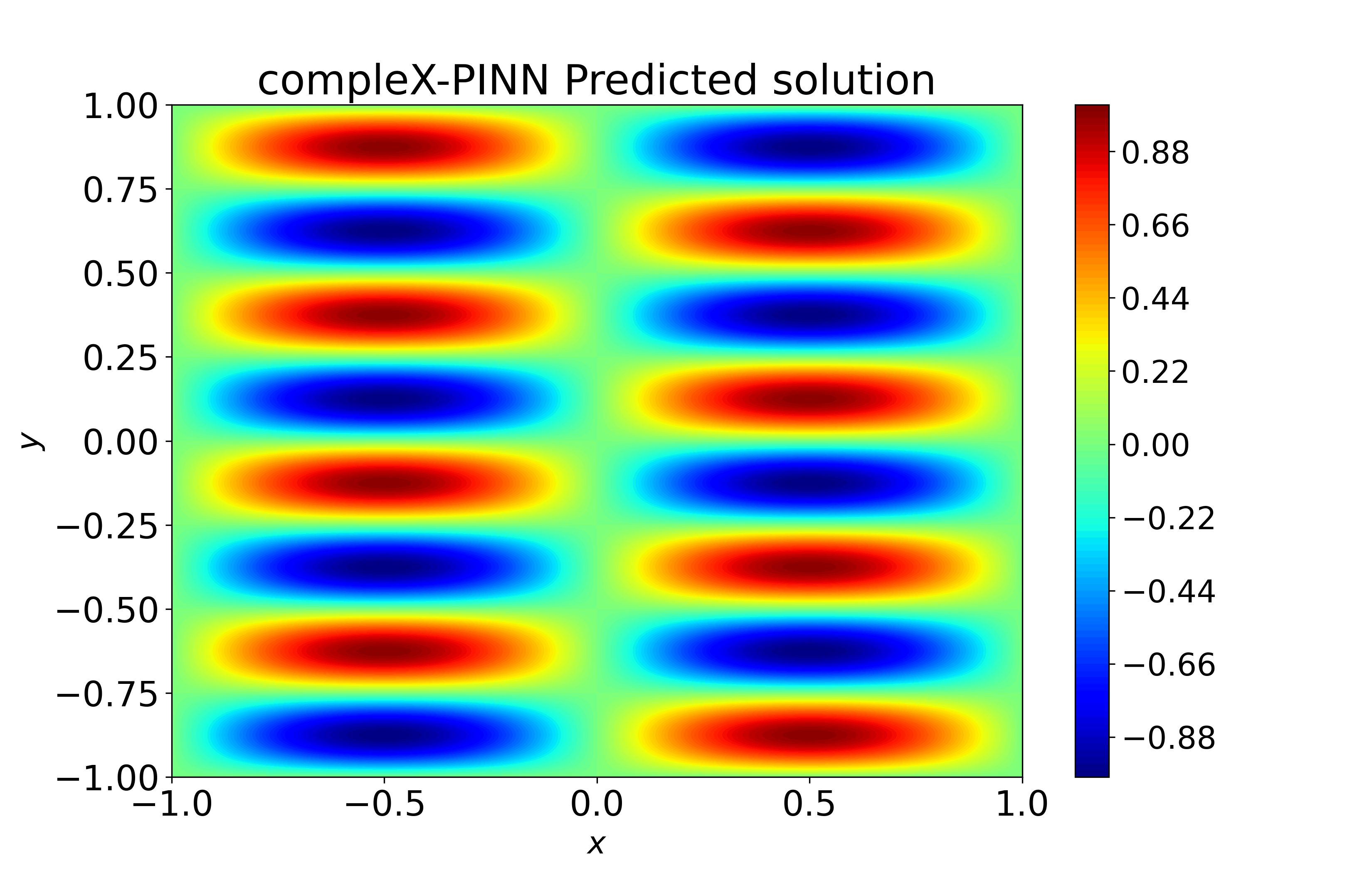}    
    \end{minipage}
    \begin{minipage}{0.45\textwidth}
    \centering
    \includegraphics[width=\linewidth]{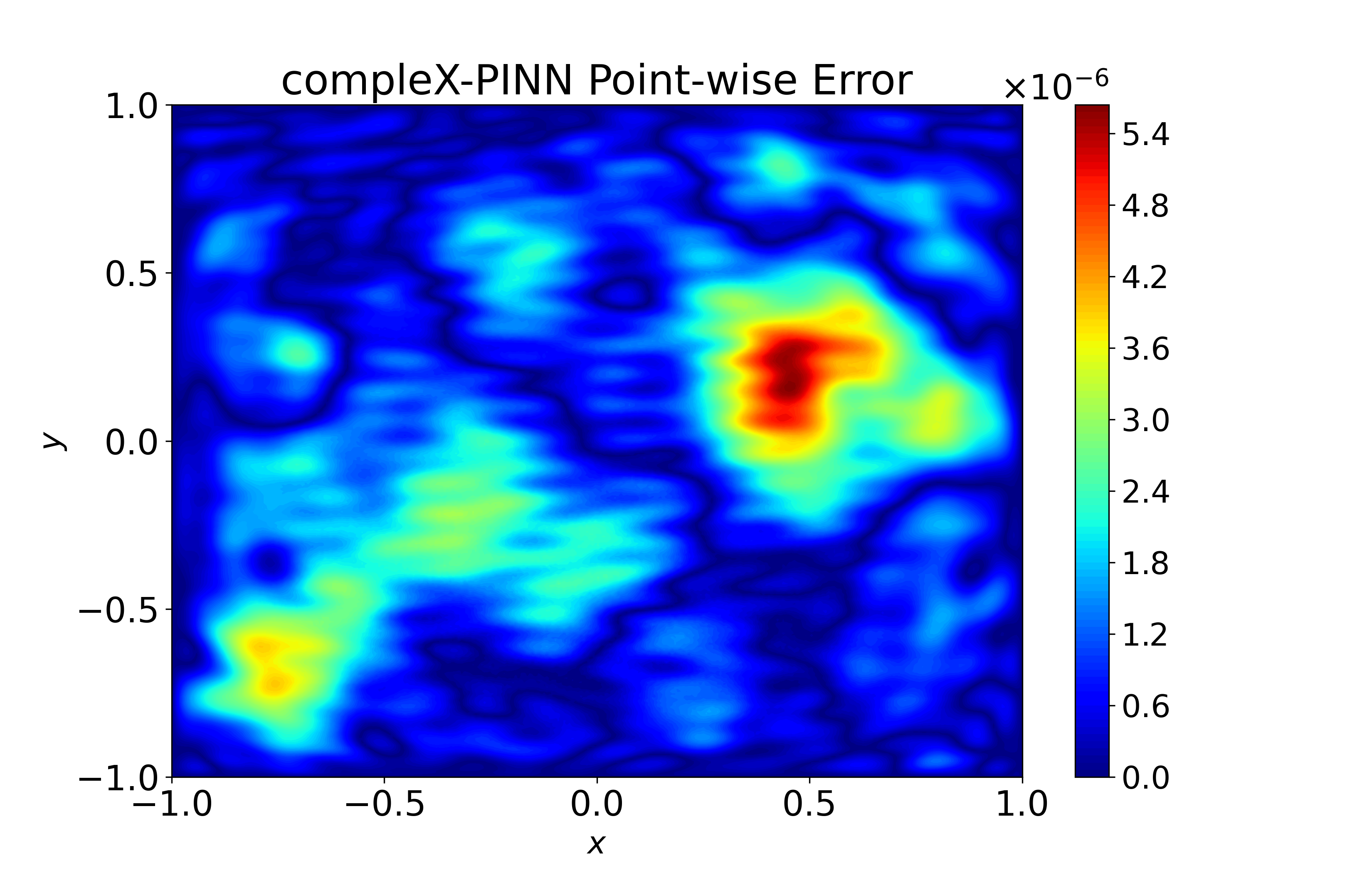}
    \end{minipage}    

    \caption{Prediction and absolute point-wise errors of compleX-PINN for the 2D Helmholtz equation. The results are obtained after fine-tuning with the L-BFGS optimizer. The final relative $L^2$ error achieved is $7.51 \times 10^{-7}$.}\label{2D Helmholtz Prediction}
\end{figure}

Given that RBA-PINN has demonstrated state-of-the-art performance on this benchmark, we compare directly with its reported results. The relative $L^2$ error history during Adam training is shown in Figure~\ref{Helmholtz L2 HIST}, where the blue curve represents the RBA-PINN output using its official code. After the Adam phase, compleX-PINN achieves a relative $L^2$ error of $2.39 \times 10^{-5}$, significantly outperforming the RBA-PINN’s corresponding error of $1.18 \times 10^{-4}$. After the L-BFGS fine-tuning phase, compleX-PINN further reduces the error to $7.51 \times 10^{-7}$, nearly one order of magnitude lower than RBA-PINN’s reported post-L-BFGS error of $5.18 \times 10^{-6}$. The final prediction results are visualized in Figure~\ref{2D Helmholtz Prediction}.

\paragraph{PIKAN Comparison (Case 4.1.1b):}\label{sec:411b}
To further validate the performance of compleX-PINN, we compare it with PIKAN~\cite{wang2025kolmogorov} using the same training protocol reported in~\cite{wang2025kolmogorov}. The PIKAN model consists of a single hidden layer with 10 neurons, the number of intervals $G = 5$, and spline order of $p = 3$. The architecture uses a multi-grid training strategy where the number of grid intervals doubles every 600 iterations. This results in a total of 750 parameters as reported in Ref.~\cite{shukla2024comprehensive}. 
Accordingly, we configure compleX-PINN with 110 neurons in the Cauchy layer, resulting in a total of 771 parameters, comparable to PIKAN. 

We use the same number of training points ($N_f = 2401$, $N_b = 200$), the same loss weights ($\lambda_F = 0.01$, $\lambda_B = 1$), and adopt the L-BFGS optimizer as done in Ref.~\cite{shukla2024comprehensive}. All models are trained for exactly 1,800 iterations. With identical training data and hyperparameters, compleX-PINN achieves a relative $L^2$ error of $6.58 \times 10^{-2}$, substantially outperforming the reported PIKAN result of $4.76 \times 10^{-1}$ by nearly an order of magnitude.

\paragraph{BsPINN Comparison (Case 4.1.1c):}\label{sec:411c}
We now evaluate compleX-PINN against BsPINN on a more challenging 2D Helmholtz problem with parameters $a_1 = a_2 = 8$, consistent with the setup in Ref.~\cite{ref34}. The BsPINN model uses a 256-16 binary configuration with five hidden layers organized hierarchically. Specifically, the first hidden layer contains one block of 256 neurons; the second layer has two blocks of 128 neurons each; the third has four blocks of 64 neurons; the fourth has eight blocks of 32 neurons; and the fifth has sixteen blocks of 16 neurons. This architecture results in 124,929 total parameters. In contrast, our compleX-PINN model reuses the architecture from Case~\ref{sec:411a} (500 neurons in the Cauchy layer), which has only 3,501 parameters—approximately 3\% of BsPINN’s size—highlighting the efficiency of our design.

\begin{figure}[!htb]
    \centering
    \begin{minipage}{0.6\textwidth}
    \centering
    \includegraphics[width=\linewidth]{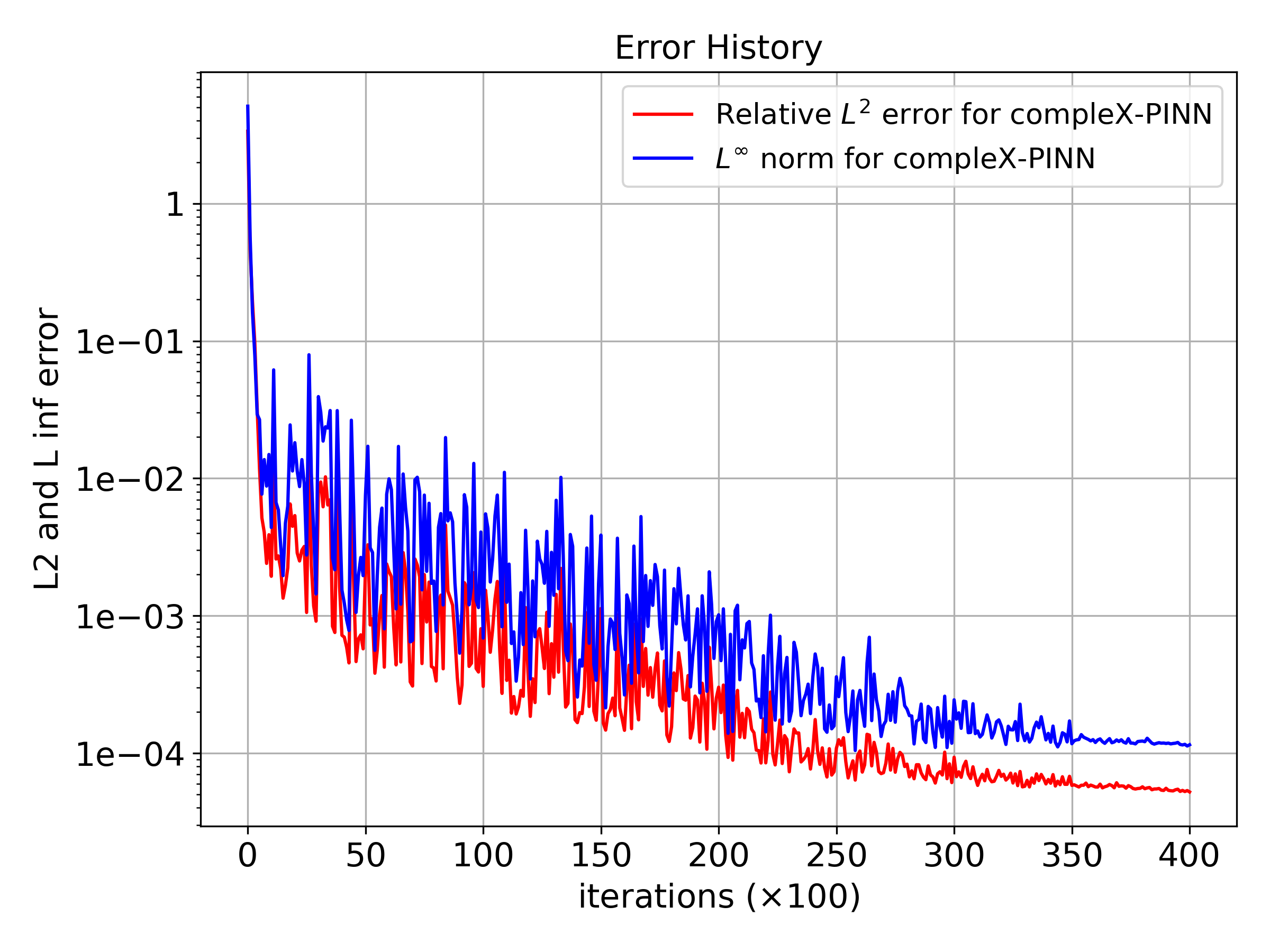}    
    \end{minipage}
    \caption{Relative $L^2$ error and $L^{\infty}$ norm history of compleX-PINN for the 2D Helmholtz equation ($a_1 = a_2 = 8$).}\label{Helmholtz L2 HIST-8}
\end{figure}
\begin{figure}[!htb]
    \centering
    \begin{minipage}{0.45\textwidth}
    \centering
    \includegraphics[width=\linewidth]{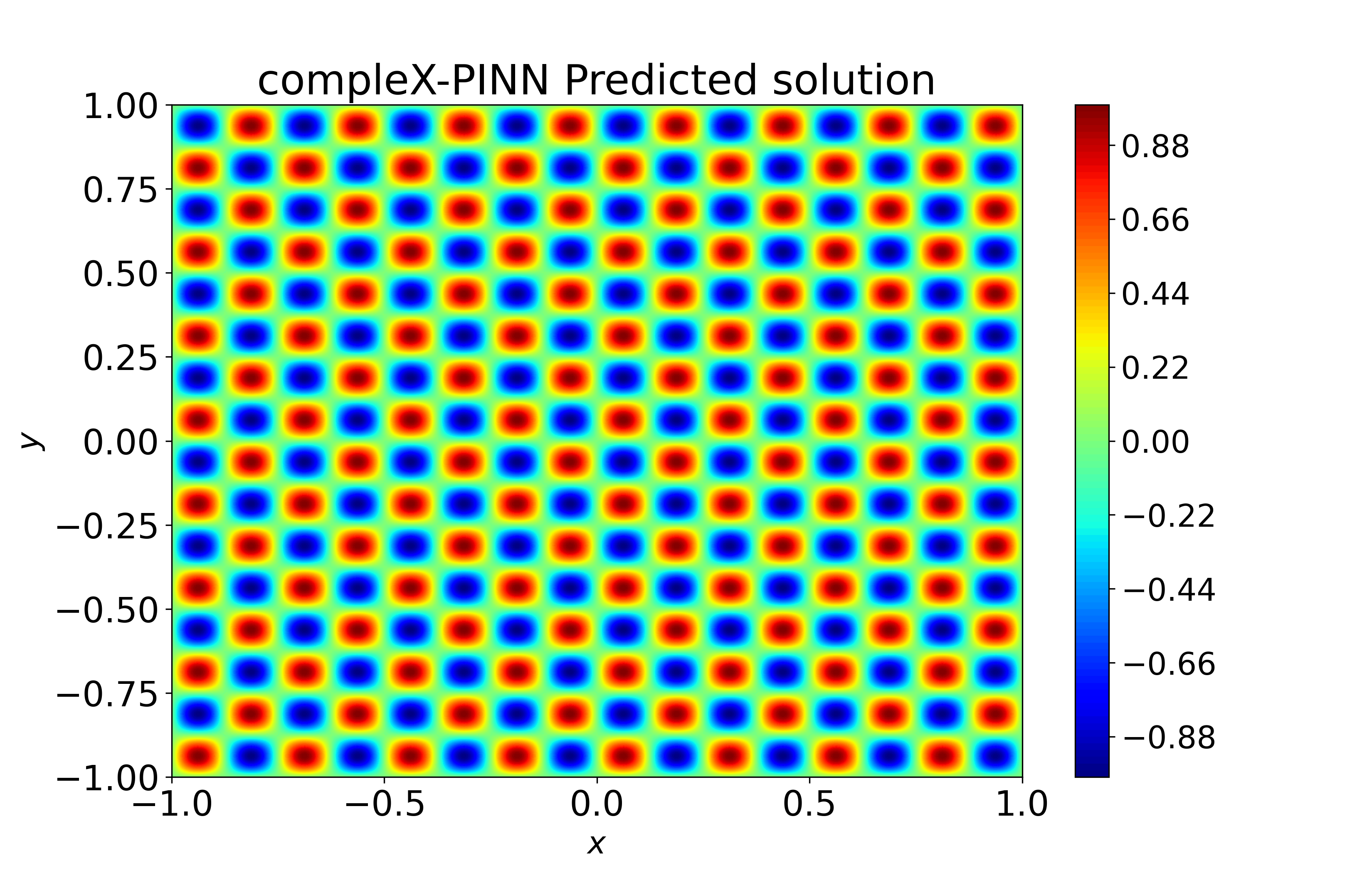}    
    \end{minipage}
    \begin{minipage}{0.45\textwidth}
    \centering
    \includegraphics[width=\linewidth]{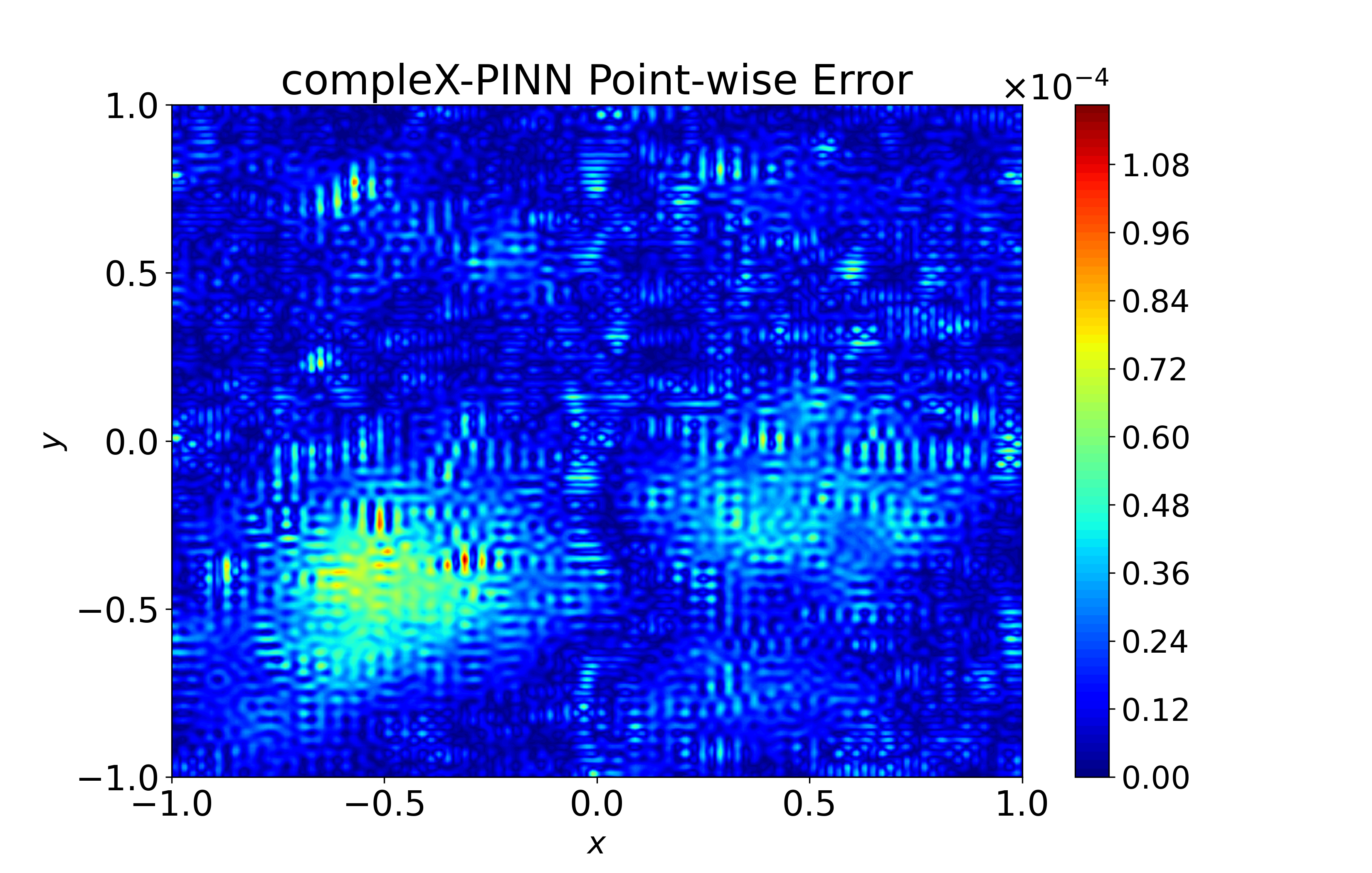}
    \end{minipage}    
    \caption{Prediction and absolute point-wise errors of compleX-PINN for the 2D Helmholtz equation when $a_1 = a_2 = 8$. The final relative $L^2$ error after training is $5.35\times 10^{-5}$.}\label{2D Helmholtz-8 Prediction}
\end{figure}

For both models, we use $N_f = 6561$ interior training points, $N_b = 320$ boundary points, and set loss weights as $\lambda_F = 1$, $\lambda_B = 100$. The models are trained using the Adam optimizer for 40,000 iterations with a learning rate of $1 \times 10^{-3}$.
Under this setup, compleX-PINN achieves a relative $L^2$ error of $5.35 \times 10^{-5}$, significantly outperforming BsPINN’s best reported result of $1.07 \times 10^{-3}$. The relative error history and heatmap of pointwise errors for compleX-PINN are shown in Figures~\ref{Helmholtz L2 HIST-8} and~\ref{2D Helmholtz-8 Prediction}, respectively.


\subsubsection{Robustness of the trainable parameters in Cauchy layers}
\label{Sec Robustness}
Recall that the Cauchy activation function introduced in~\eqref{(12)} contains trainable parameters ${\mu_1, \mu_2, d}$, which are manually initialized. By default, we initialize each of these parameters to 0.1. However, it is important to investigate whether this initialization impacts model performance.


In this section, we examine the robustness of these parameters by varying their initial values among $\{0.01, 0.5, 1.0\}$. Testing all possible combinations of $\{\mu_1, \mu_2, d\}$ would be computationally intensive. To balance thoroughness with feasibility, we adopt a controlled perturbation strategy: we fix two parameters at their default value (0.1) and vary the third. This allows us to isolate the effect of each individual parameter’s initialization on the model’s performance. 

All experiments follow the same training configuration described in Case~\ref{sec:411a}a, and we restrict our analysis to the Adam training phase.


\begin{table}[ht]
\centering
\caption{Robustness of model performance with respect to the initialization of Cauchy activation function parameters ${\mu_1, \mu_2, d}$. Each parameter is varied individually while the other two are fixed at 0.1. Reported values are the relative $L^2$ errors after 20,000 Adam training iterations (mean ± standard deviation over 5 trials). "NaN" indicates numerical instability during training.}
\label{tab:parameter_robustness}
\begin{tabular}{lccc}
\toprule
\multirow{2}{*}{Parameter} & \multicolumn{3}{c}{Initialization Value} \\
\cmidrule{2-4}
 & 0.01 & 0.5 & 1.0 \\
\midrule
$\mu_1$ (fixed: $\mu_2=0.1, d=0.1$) & $(2.38\pm 0.18)\times 10^{-5}$ & $(2.39\pm 0.47)\times 10^{-5}$  & $(2.58\pm 0.41)\times 10^{-5}$ \\

$\mu_2$ (fixed: $\mu_1=0.1, d=0.1$) & $(2.69\pm 0.22)\times 10^{-5}$ & $(2.99\pm 0.57)\times 10^{-5}$ & $(3.47\pm 0.83)\times 10^{-5}$ \\

$d$ (fixed: $\mu_1=0.1, \mu_2=0.1$) & NaN & $(3.38\pm 0.46)\times 10^{-5}$ &  $(7.85\pm 0.16)\times 10^{-5}$ \\
\bottomrule
\end{tabular}
\end{table}

As shown in Table~\ref{tab:parameter_robustness}, variations in the initial values of $\mu_1$ and $\mu_2$ result in relatively consistent model performance, with only minor fluctuations in relative $L^2$ error. This suggests that the model is robust to the initialization of these two parameters. In contrast, the initialization of $d$ has a more pronounced impact. Specifically, setting $d = 0.01$ leads to numerical instability (NaN results), while larger values increase the relative error. This sensitivity is expected, as $d$ controls the "width" of the Cauchy activation function. As a scale parameter, small values of $d$ sharply localize the activation, potentially leading to gradient instability and poor generalization. These findings highlight the importance of avoiding near-zero initializations for $d$ to ensure stable and effective training.


\subsection{3D Helmholtz equation}
\label{Sec 3D Helmholtz}
We now extend our investigation to the 3D Helmholtz equation. Specifically, we consider the following problem:
\begin{align*}
    &\Delta u + k^2u = q(x,y,z) , \quad (x,y,z) \in \Omega, \\
    &u(x,y,z) = 0, \quad (x,y,z)\in\partial\Omega, 
\end{align*}
where $k = 1$, and the analytical solution is given by
\begin{align*}
    u(x,y,z) = \sin(2\pi x)\sin(2\pi y)\sin(2\pi z),
\end{align*}
with the source term $q(x, y, z)$ inferred accordingly.

All models incorporate a hard constraint for the boundary condition, expressed as $\hat{u} = (1 - x^2)(1 - y^2)(1 - z^2)\hat{u}_{\mathcal{NN}}$, and use $N_f = 10{,}000$ residual training points. Training is performed using the Adam optimizer for 20k iterations with the same learning rate and exponential decay scheduler as described in Case~\ref{sec:411a}a. 

The compleX-PINN employs 500 neurons in the Cauchy layer. For baseline models, hyperparameter tuning is conducted as described at the beginning of Section~\ref{Sec4}, and the best-performing configurations are selected accordingly: RBA-PINN uses $D_{\text{RBA}} = 4$ and $N_{\text{RBA}} = 50$; PIKAN uses $G = 4$ and $N_{\text{KAN}} = 5$; BsPINN adopts a 5-layer hierarchical architecture with block sizes halving from 256 to 16, resulting in a 256–16 configuration.

Table~\ref{tab:3D Helmholtz time} summarizes the number of trainable parameters and the GPU time per 100 training iterations for each model. Notably, compleX-PINN and PIKAN have significantly fewer parameters than RBA-PINN and BsPINN, highlighting their advantage in model compactness and memory efficiency. While compleX-PINN has more parameters than PIKAN, it requires substantially less training time per 100 iterations, demonstrating superior computational efficiency.

\begin{table}[!ht]
\centering
\caption{Number of trainable parameters and GPU training time per 100 iterations for different models. This comparison highlights the computational and memory efficiency of compleX-PINN relative to other baseline methods.}
\label{tab:3D Helmholtz time}
\begin{tabular}{lrr}
\toprule
Model & Number of Parameters & GPU Time (s/100 iters) \\
\midrule
compleX-PINN & 4,001 & 5.23 \\
RBA-PINN     & 7,901 & 3.05 \\
BsPINN       & 125,185 & 4.81 \\
PIKAN        & 405 & 11.70 \\
\bottomrule
\end{tabular}
\end{table}

Table~\ref{tab:3D Helmholtz errors} reports the numerical performance under two evaluation criteria: (i) the same number of training iterations (20k), and (ii) equivalent total training time. After 20k iterations, compleX-PINN achieves a relative $L^2$ error of $5.54 \times 10^{-4}$. PIKAN has the second smallest $L^2$ error of $7.27 \times 10^{-4}$ after 20k iterations, despite its longer training time. When training time is equalized, BsPINN and RBA-PINN attain relative $L^2$ errors of $9.94 \times 10^{-4}$ and $1.03 \times 10^{-3}$, respectively. Overall, compleX-PINN consistently outperforms other methods and yields the best performance across both evaluation metrics, even when trained under time constraints.

\begin{table}[!ht]
\centering
\caption{Relative $L^2$ errors and $L^{\infty}$ norms for each model under two evaluation settings—fixed number of iterations (20k) and equivalent computational time.}
\label{tab:3D Helmholtz errors}
\begin{tabular}{lcccc}
\toprule
\multirow{2}{*}{\makecell{Model\\(total 20k iter.)}} & \multicolumn{2}{c}{Relative $L^2$ Norm} & \multicolumn{2}{c}{$L^\infty$ Norm} \\
\cmidrule(lr){2-3} \cmidrule(lr){4-5} 
& Same Iter. & Same Comput. Time  & Same Iter. & Same Comput. Time  \\
\midrule
compleX-PINN & \multicolumn{2}{c}{\bm{$5.54 \times 10^{-4}$}} & \multicolumn{2}{c}{\bm{$9.27 \times 10^{-4}$}} \\
RBA-PINN     & $2.71 \times 10^{-3}$ & $1.03 \times 10^{-3}$ & $3.13 \times 10^{-2}$ & $3.01 \times 10^{-2}$ \\
BsPINN       & $1.03 \times 10^{-3}$ & $9.94 \times 10^{-4}$ & $4.25 \times 10^{-3}$ & $4.18 \times 10^{-3}$ \\
PIKAN        & $7.27 \times 10^{-4}$ & $3.58 \times 10^{-3}$ & $2.72 \times 10^{-3}$ & $5.07 \times 10^{-3}$ \\
\bottomrule
\end{tabular}
\end{table}



\subsection{3D heat equation}
\label{Sec 3D Heat}
We now evaluate model performance on the 3D heat equation given by:
\begin{align}
    &\Delta u(\mathbf{x},t) = 12\pi^2  u_t , \quad \mathbf{x} \in \Omega:=[0,1]^3, t\in T:=[0,10], \\
    &u(\mathbf{x},t) = 0, \quad \mathbf{x}\in\partial\Omega, t\in T, \\
    &u(\mathbf{x},0) = \sin(2\pi x_1)\sin(2\pi x_2)\sin(2\pi x_3), \quad \mathbf{x} \in \Omega.
\end{align}
The exact solution is:
\begin{align}
    u(\mathbf{x}, t) = e^{-t}\sin(2\pi x_1)\sin(2\pi x_2)\sin(2\pi x_3).
\end{align}
To enforce initial and boundary conditions, we apply a hard constraint of the form:
\[\hat{u} = t \prod_{i=1}^3x_i(1-x_i) \cdot \hat{u}_{\mathcal{NN}}  + \prod_{i=1}^3\sin(2\pi x_i).\]
We use $N_f = 10,000$ residual training points across space and time.
The compleX-PINN is configured with 300 neurons in its Cauchy layer. For comparison, the baseline models are optimized with the following architectures based on prior tuning: RBA-PINN uses 4 hidden layers with 80 neurons each; BsPINN adopts a 256-16 block-wise configuration; and PIKAN uses $N_{\text{KAN}} = 10$ and $G = 2$.

\begin{figure}[!htb]
    \centering
    \begin{minipage}{0.45\textwidth}
    \centering
    \includegraphics[width=\linewidth]{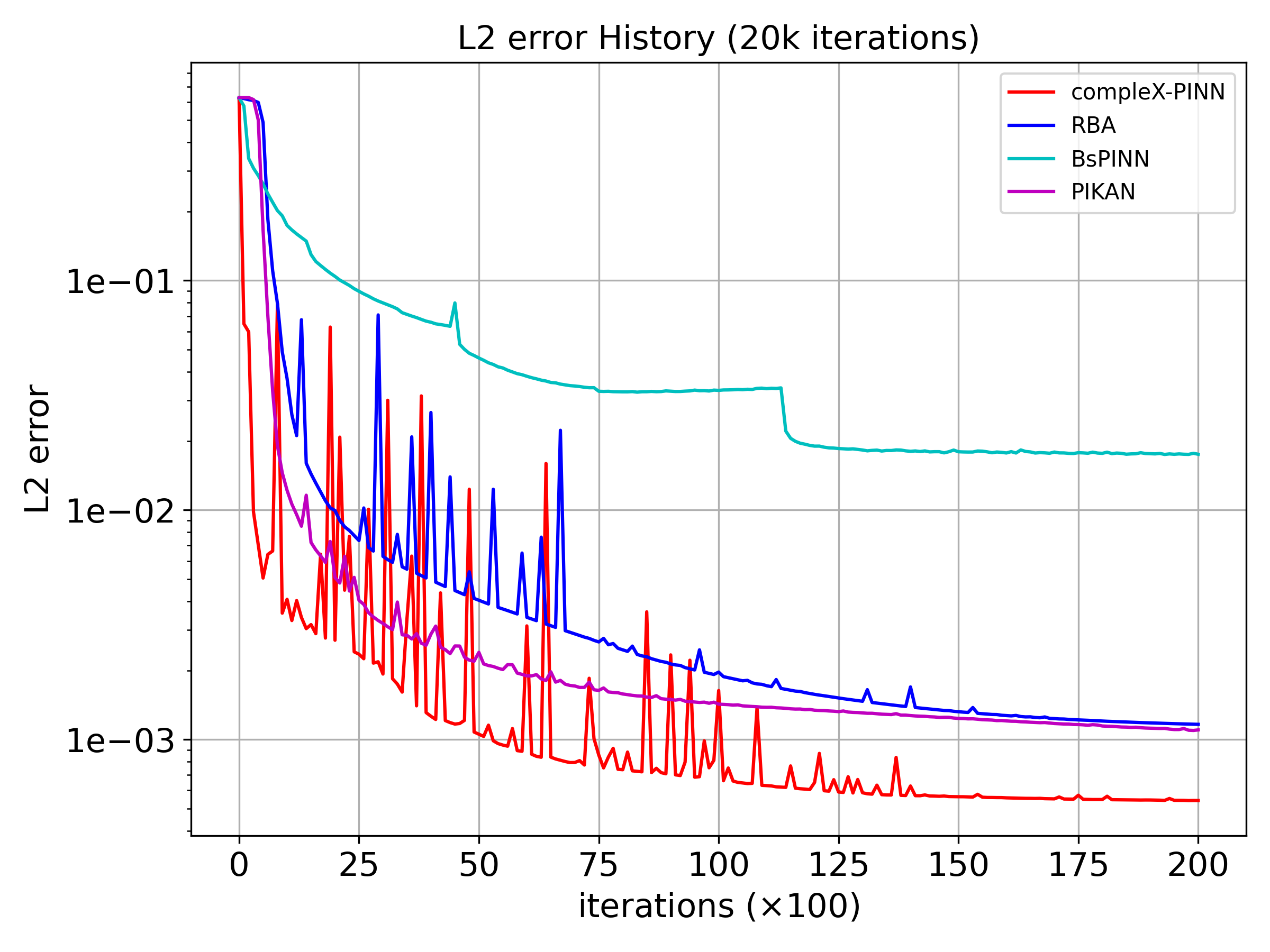}    
    \end{minipage}
    \begin{minipage}{0.45\textwidth}
    \centering
    \includegraphics[width=\linewidth]{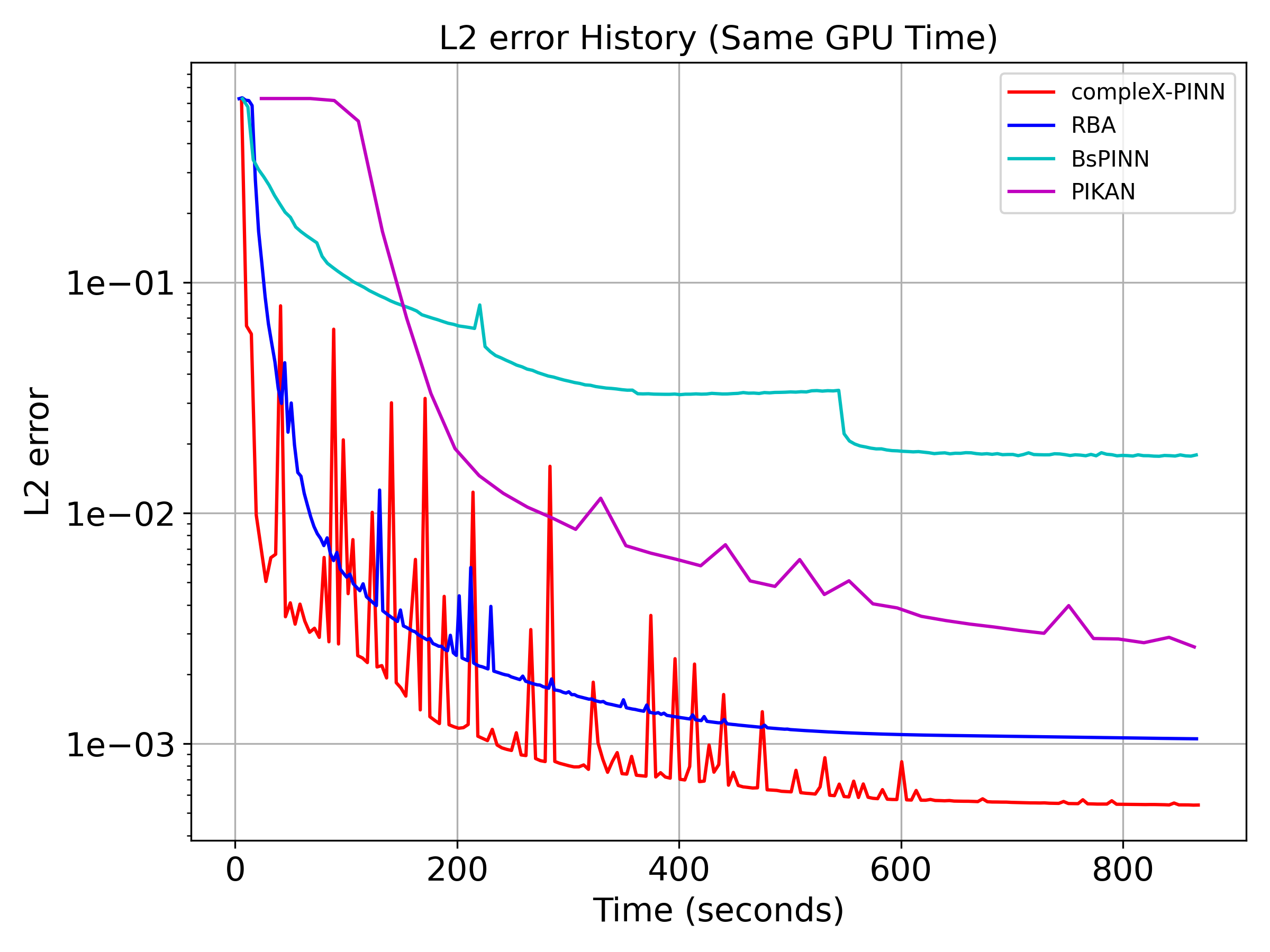}
    \end{minipage}    

    \caption{Relative $L^2$ error history of compleX-PINN (red), RBA-PINN (blue), BsPINN (cyan), and PIKAN (magenta) on the 3D heat equation. The left plot shows error evolution over 20,000 training iterations. The right plot presents the error history when models are trained for the same total GPU time as compleX-PINN.}\label{3D Heat HIST}
\end{figure}

All models are trained using the Adam optimizer with an initial learning rate of $5 \times 10^{-3}$, decayed exponentially by a factor of 0.85 every 1,000 iterations, until below $1 \times 10^{-5}$. Figure~\ref{3D Heat HIST} presents the training history of the relative $L^2$ error, and Table~\ref{tab:3D Heat errors} summarizes each model's performance under two evaluation protocols: fixed number of training iterations (20,000) and fixed total training time (matched to compleX-PINN).

\begin{table}[!ht]
\centering
\caption{3D heat equation: Relative $L^2$ errors and $L^{\infty}$ norms for each model under two evaluation settings—fixed number of iterations (20k) and equivalent computational time.}
\label{tab:3D Heat errors}
\begin{tabular}{lccccc}
\toprule
\multirow{2}{*}{\makecell{Model\\(total 20k iter.)}} & \multicolumn{2}{c}{Relative $L^2$ Norm} & \multicolumn{2}{c}{$L^\infty$ Norm} & \multirow{2}{*}{\makecell{GPU Time\\(per 100 iter.)}} \\
\cmidrule(lr){2-3} \cmidrule(lr){4-5} 
& Same Iter. & Same Comput. Time & Same Iter. & Same Comput. Time &  \\
\midrule
compleX-PINN & \multicolumn{2}{c}{\bm{$2.73 \times 10^{-4}$}} & \multicolumn{2}{c}{\bm{$7.67 \times 10^{-4}$}} & \text{4.25s} \\
RBA-PINN  & $1.16 \times 10^{-3}$  & $1.06 \times 10^{-3}$ & $2.27 \times 10^{-3}$  & $2.23 \times 10^{-3}$ & \text{3.03s} \\
BsPINN  & $1.75 \times 10^{-2}$ & $1.77 \times 10^{-2}$ & $3.58 \times 10^{-2}$ & $3.76 \times 10^{-2}$ & \text{5.05s} \\
PIKAN   & $1.10 \times 10^{-3}$ & $3.12 \times 10^{-3}$ & $1.39 \times 10^{-3}$ & $4.19 \times 10^{-3}$ & \text{20.8s} \\
\bottomrule
\end{tabular}
\end{table}    

Under identical iteration count, compleX-PINN achieves a relative $L^2$ error of $2.73 \times 10^{-4}$, significantly outperforming the other three methods. This advantage persists even under equal runtime. For example, although RBA-PINN completes over 28,000 iterations within the same time (due to its faster per-iteration speed of 3.03s per 100 iterations), its final error remains at $1.06 \times 10^{-3}$, still nearly 4× higher than compleX-PINN. PIKAN, despite achieving reasonable accuracy ($1.10 \times 10^{-3}$) with 20k iterations, suffers from high computational cost (20.8s per 100 iterations), allowing only ~4,100 iterations under equal runtime. Consequently, its error increases to $3.12 \times 10^{-3}$, over 11× higher than compleX-PINN. 

In terms of maximum point-wise error, compleX-PINN again demonstrates superiority, achieving the lowest $L^\infty$ norm of $7.67 \times 10^{-4}$, underscoring its effectiveness in reducing localized spatiotemporal errors. These results highlight the strength of the Cauchy-layer-based architecture in enhancing both global and worst-case accuracy.

\subsection{High-dimensional Poisson equation}
\label{Section Poisson}
We evaluate the performance of compleX-PINN on high-dimensional problems using the Poisson equation in five and ten dimensions. The governing equation is defined on the hypercube domain $\Omega:= [-1,1]^d$, with $d \in {5, 10}$:
\begin{align}
    -\Delta u &= f \quad\text{in }\Omega,\\
    u &= g \quad\text{on }\partial\Omega,
\end{align}
where the exact solution is given by $u(\mathbf{x}) = \sum_{i=1}^d \sin(\pi x_i)$ and the corresponding source term is $f(\mathbf{x}) = \pi^2 \sum_{i=1}^d \sin(\pi x_i)$. This setup provides a controlled and scalable benchmark for assessing the accuracy and efficiency of the models.

\begin{figure}[!htb]
    \centering
    \begin{minipage}{0.45\textwidth}
    \centering
    \includegraphics[width=\linewidth]{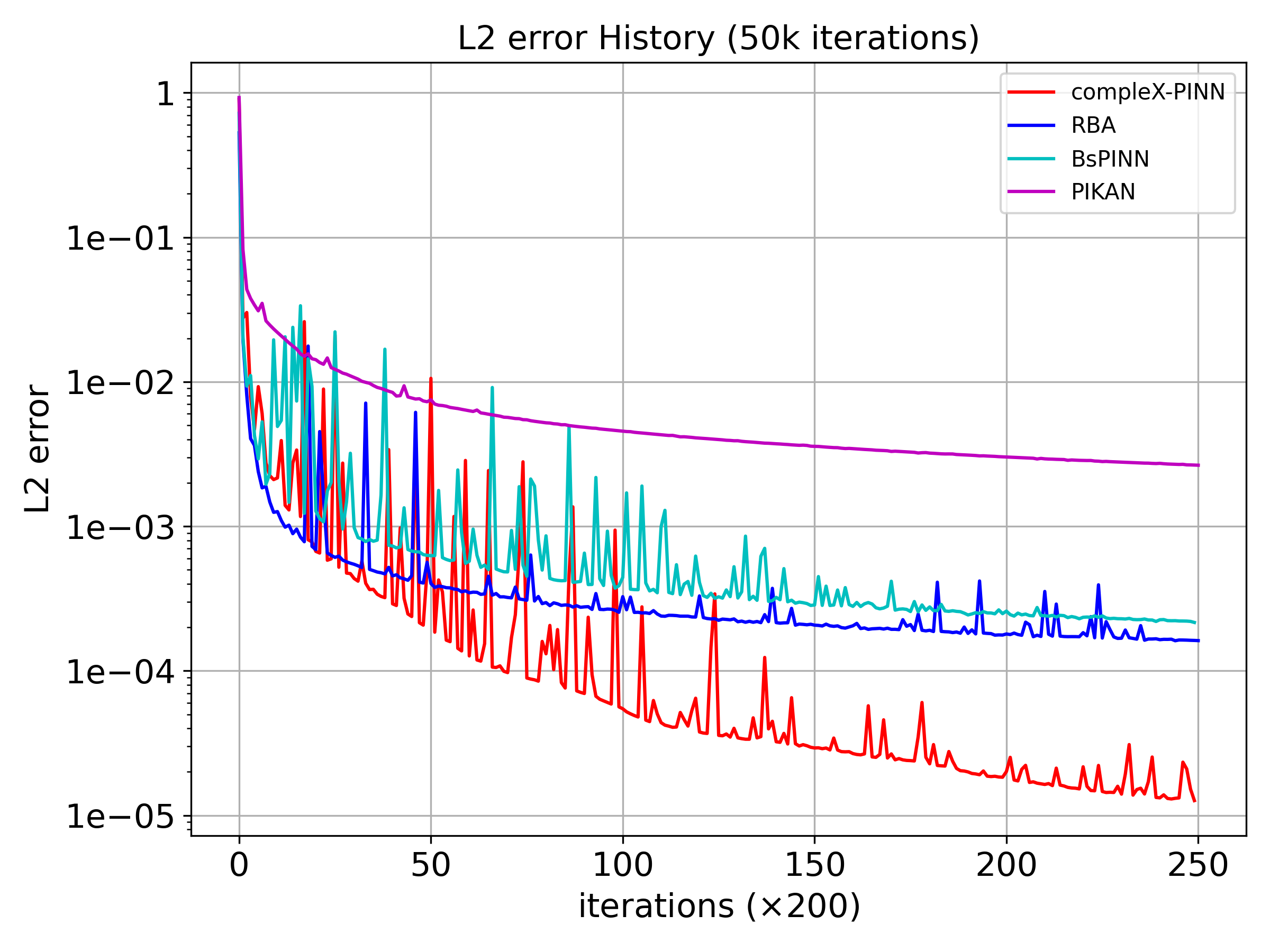}    
    \end{minipage}
    \begin{minipage}{0.45\textwidth}
    \centering
    \includegraphics[width=\linewidth]{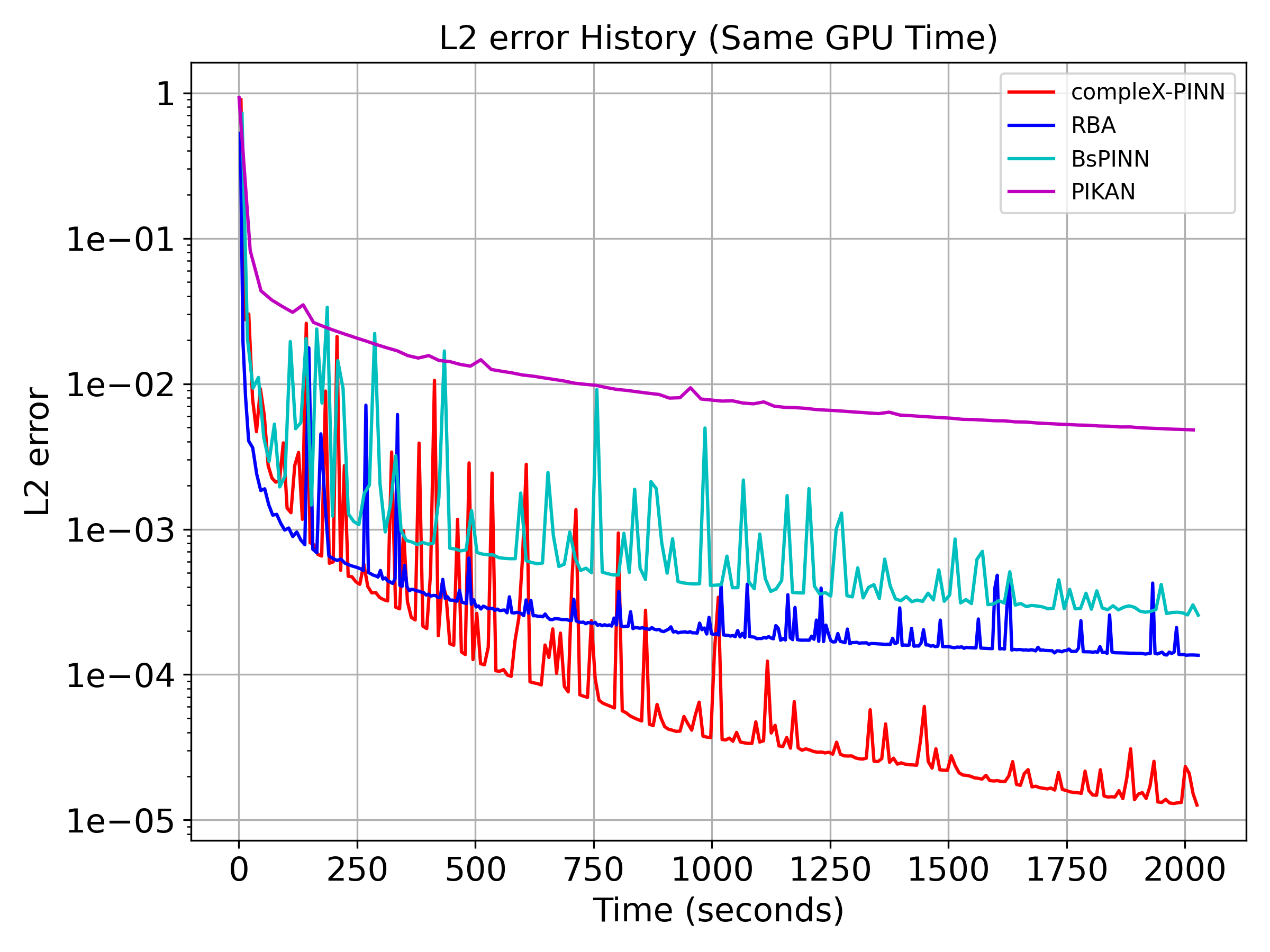}
    \end{minipage}    

    \begin{minipage}{0.45\textwidth}
    \centering
    \includegraphics[width=\linewidth]{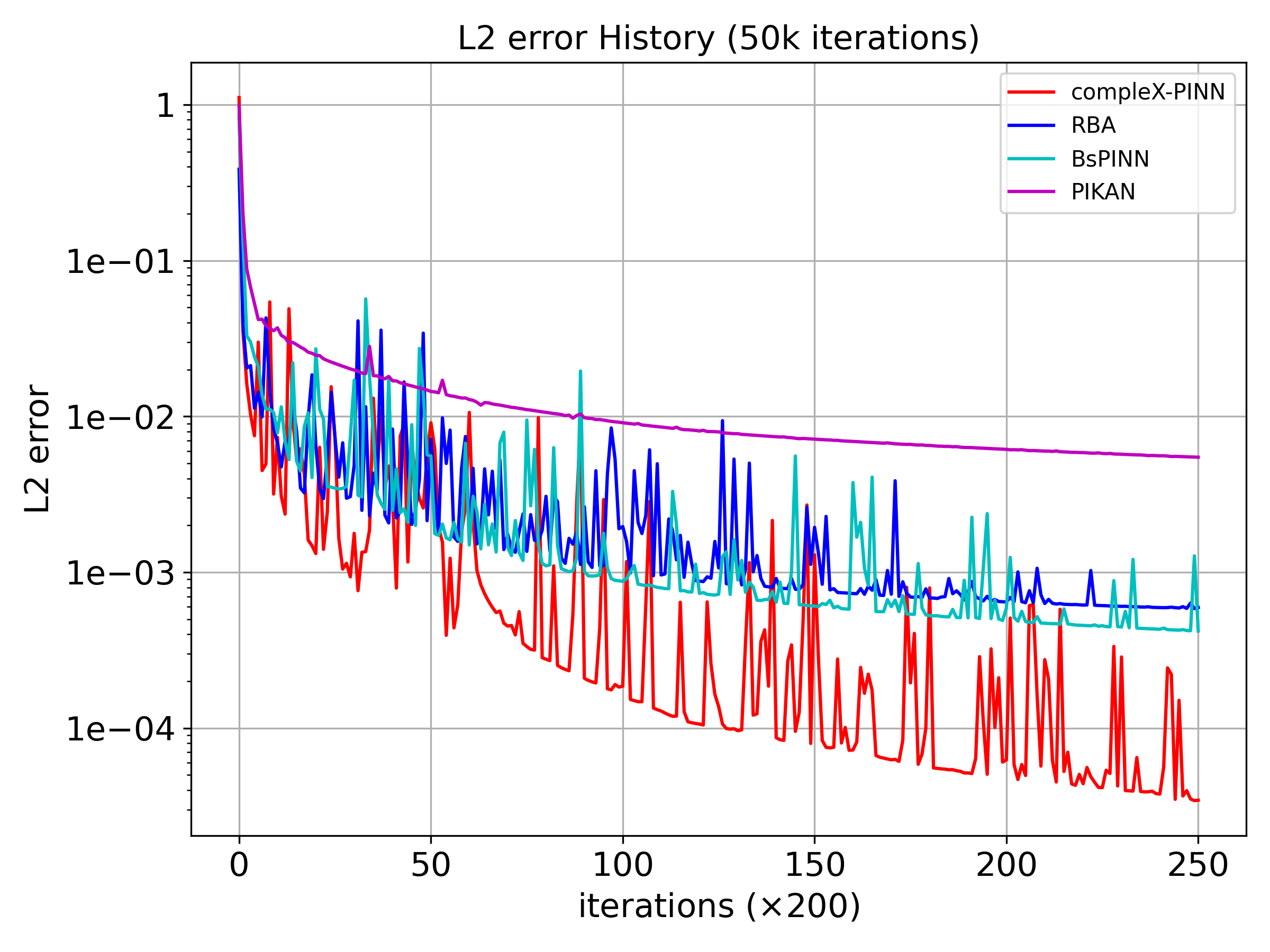}    
    \end{minipage}
    \begin{minipage}{0.45\textwidth}
    \centering
    \includegraphics[width=\linewidth]{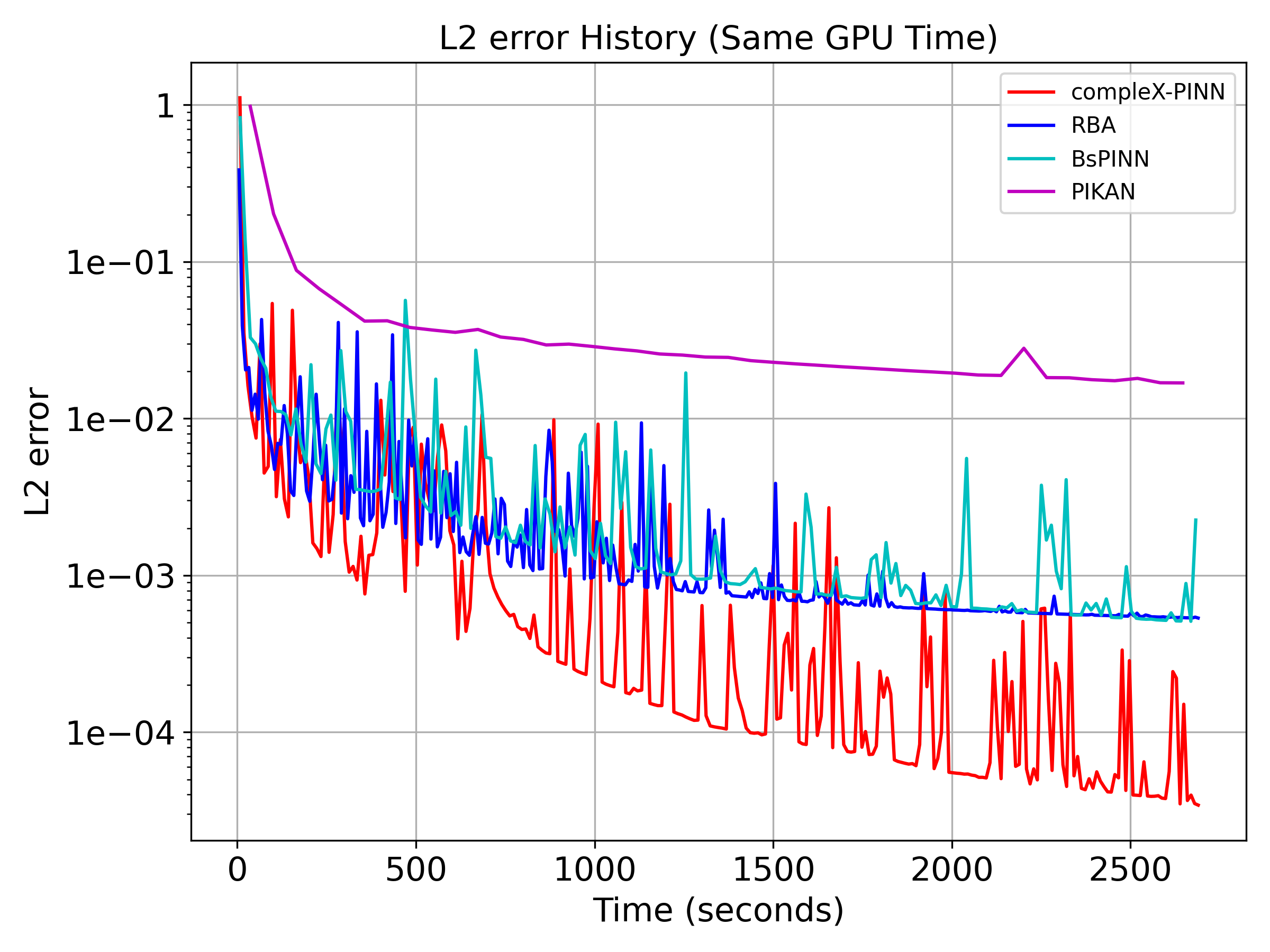}
    \end{minipage} 

    \caption{Relative $L^2$ error history of compleX-PINN (red), RBA-PINN (blue), BsPINN (cyan), and PIKAN (magenta) for the Poisson equation in 5D (top row) and 10D (bottom row). The left column presents error trajectories over 50,000 training iterations, while the right column compares performance under equal GPU time budgets. compleX-PINN consistently demonstrates faster convergence and superior accuracy across both dimensional settings.}\label{fig:Poisson HIST}
\end{figure}

For both the 5D and 10D cases, all models are trained under identical settings: $N_f = 10,000$ interior collocation points, $N_b = 500$ boundary points, and 50,000 training iterations using the Adam optimizer with an initial learning rate of $1 \times 10^{-2}$. A learning rate scheduler with decay factor 0.85 and decay step 1,000 is applied, and the loss weights are fixed at $\lambda_F = 1$ and $\lambda_B = 100$. The compleX-PINN employs 200 neurons in the Cauchy layer. Baseline models are configured with their optimal architectures: RBA-PINN uses 3 hidden layers of 200 neurons; PIKAN adopts $N_{\text{KAN}} = 15$ with number of intervals $G = 9$; and BsPINN follows a 128-8 architecture. The results are summarized in Table~\ref{tab:Poisson errors} and visualized in Figure~\ref{fig:Poisson HIST}.

In the 5D case, compleX-PINN achieves a relative $L^2$ error of $1.66 \times 10^{-5}$ and an $L^{\infty}$ norm of $4.09 \times 10^{-3}$, significantly outperforming all baselines by one to two orders of magnitude. When extended to 10D, compleX-PINN maintains remarkable robustness, with its $L^2$ error increasing only slightly to $3.44 \times 10^{-5}$ (a 2.1× growth) and its $L^{\infty}$ norm rising to $4.54 \times 10^{-3}$ (1.1× growth). In contrast, baseline models experience much greater degradation from 5D to 10D. For instance, RBA-PINN’s $L^{\infty}$ norm increases by 4.1× (from $1.58 \times 10^{-2}$ to $6.45 \times 10^{-2}$), BsPINN’s by 5.9× (from $1.09 \times 10^{-2}$ to $6.46 \times 10^{-2}$), and PIKAN’s nearly doubles (from $1.57 \times 10^{-1}$ to $3.29 \times 10^{-1}$). These results underscore compleX-PINN’s ability to suppress error amplification in high-dimensional settings.


In terms of efficiency, compleX-PINN also demonstrates strong scalability. The GPU time per 100 iterations increases by only 33\% from 4.05s (5D) to 5.38s (10D)—notably lower than the growth observed in RBA-PINN (45\%), BsPINN (47\%), and PIKAN (58\%). 
These results demonstrate that compleX-PINN not only achieves high accuracy in high dimensions but also maintains strong computational efficiency and stability, effectively overcoming the key challenges of high-dimensional PDE solvers.


\begin{table}[ht]
\centering
\caption{5D and 10D Poisson equations: Relative $L^2$ errors and $L^{\infty}$ norms for each model under two evaluation settings—fixed number of iterations (20k) and equivalent computational time.}
\label{tab:Poisson errors}

\begin{tabular}{lccccc}
\toprule
\multirow{2}{*}{\makecell{Model (5D)\\(total 50k iter.)}} & \multicolumn{2}{c}{Relative $L^2$ Norm} & \multicolumn{2}{c}{$L^\infty$ Norm} & \multirow{2}{*}{\makecell{GPU Time\\(per 100 iter)}} \\
\cmidrule(lr){2-3} \cmidrule(lr){4-5} 
& Same Iter. & Same Comput. Time & Same Iter. & Same Comput. Time &  \\
\midrule
compleX-PINN & \multicolumn{2}{c}{\bm{$1.66 \times 10^{-5}$}} & \multicolumn{2}{c}{\bm{$4.09 \times 10^{-3}$}} & \text{4.05s} \\
RBA-PINN  & $2.77 \times 10^{-4}$ & $2.42 \times 10^{-4}$  & $1.58 \times 10^{-2}$ & $1.40\times 10^{-2}$ & \text{2.86s} \\
BsPINN  & $2.19 \times 10^{-4}$ & $2.49\times 10^{-4}$ & $1.09 \times 10^{-2}$ & $1.14 \times 10^{-2}$ & \text{5.11s} \\
PIKAN   & $2.64\times 10^{-3}$ & $7.37\times 10^{-3}$ & $1.57\times 10^{-1}$ & $4.18\times 10^{-1}$& \text{21.2s} \\
\bottomrule
\end{tabular}

\vspace{0.5em} 

\begin{tabular}{lccccc}
\toprule
\multirow{2}{*}{\makecell{Model (10D)\\(total 50k iter.)}} & \multicolumn{2}{c}{Relative $L^2$ Norm} & \multicolumn{2}{c}{$L^\infty$ Norm} & \multirow{2}{*}{\makecell{GPU Time\\(per 100 iter)}} \\
\cmidrule(lr){2-3} \cmidrule(lr){4-5} 
& Same Iter. & Same Comput. Time & Same Iter. & Same Comput. Time &  \\
\midrule
compleX-PINN & \multicolumn{2}{c}{\bm{$3.44\times 10^{-5}$}} & \multicolumn{2}{c}{\bm{$4.54\times 10^{-3}$}} & \text{5.38s} \\
RBA-PINN  & $5.84\times 10^{-4}$  & $5.29\times 10^{-4}$ & $6.45\times 10^{-2}$ & $6.83\times 10^{-2}$ & \text{4.16s} \\
BsPINN  & $4.19\times 10^{-4}$  & $5.30\times 10^{-4}$ & $6.46\times 10^{-2}$ & $6.93\times 10^{-2}$& \text{7.52s} \\
PIKAN   & $5.48\times 10^{-3}$ & $1.67\times 10^{-2}$ & $3.29\times 10^{-1}$ & $8.48\times 10^{-1}$ &\text{33.6s} \\
\bottomrule
\end{tabular}

\end{table}

\section{Conclusion and Future Work}
\label{Sec discussion}
We propose compleX-PINN, a novel single-layer physics-informed neural network architecture grounded in the Cauchy integral formula. 
Theoretically, this framework reduces parameter complexity while maintaining rigorous approximation guarantees, directly addressing the inefficiencies and depth-related limitations of conventional PINNs. 
Empirically, compleX-PINN achieves state-of-the-art accuracy and stability across a variety of high-dimensional PDE benchmarks, consistently delivering orders-of-magnitude lower relative $L^2$ and $L^\infty$ errors than leading alternatives such as RBA-PINN, BsPINN, and PIKAN. 
By uniting mathematical rigor with practical scalability, compleX-PINN offers a lightweight yet powerful paradigm for physics-informed machine learning.
 


Looking ahead, a key direction is the extension of compleX-PINN to non-smooth and discontinuous systems, such as PDEs with shock waves or sharp interfaces. While the current framework excels in smooth or analytic settings, its reliance on the analyticity assumptions inherent in the Cauchy integral formula may limit its direct applicability to such problems. This limitation indicates a potential future research direction. We will conduct deeper research in this area to explore effective solutions. Furthermore, deep or hierarchical variants of compleX-PINN—such as networks composed of multiple Cauchy layers—could enhance its expressiveness for multiscale phenomena without compromising training efficiency. Finally, integrating compleX-PINN with operator learning frameworks presents an exciting opportunity for emulating parametric systems efficiently, bridging the gap between classical numerical methods and modern data-driven approaches.

\section*{Acknowledgments}
This work was partially supported by the National Natural Science Foundation of China (72495131, 82441027), Guangdong Provincial Key Laboratory of Mathematical Foundations for Artificial Intelligence (2023B1212010001), Shenzhen Stability Science Program, and the Shenzhen Science and Technology Program under grant no. ZDSYS20211021111415025.

\bibliographystyle{unsrt}  
\bibliography{ref}
\end{document}